%% file: main.tex
\documentclass[11pt,a4paper]{article}
\usepackage[hyperref]{acl2021}
\usepackage{times}
\usepackage{latexsym}

\usepackage{microtype}
\usepackage{array}
\usepackage{pifont}
\usepackage{tabularx}
\usepackage{adjustbox}
\usepackage{multirow}
\usepackage{enumitem}
\usepackage{xspace}
\usepackage{tcolorbox}
\usepackage{booktabs,subcaption,amsfonts,dcolumn}
\usepackage{hyperref}
\usepackage{url}
\usepackage{amsmath,amsthm,amsfonts,amssymb,bm,stmaryrd}
\usepackage{xcolor}		% make links dark blue
\usepackage[noorphans,vskip=0.75ex,leftmargin=2ex]{quoting}

\aclfinalcopy % Uncomment this line for the final submission
\def\aclpaperid{982} %  Enter the acl Paper ID here

\definecolor{darkblue}{rgb}{0, 0, 0.5}
\hypersetup{colorlinks=true, citecolor=darkblue, linkcolor=darkblue, urlcolor=darkblue}

\usepackage[compact]{titlesec}
\titlespacing{\section}{0pt}{2ex}{1ex}
\titlespacing{\subsection}{0pt}{1ex}{1ex}

\input{header}

\title{Making Pre-trained Language Models Better Few-shot Learners}

\author{Tianyu Gao$^{\dagger*}$ \quad Adam Fisch$^{\ddagger*}$ \quad Danqi Chen$^{\dagger}$ \\
$^{\dagger}$Princeton University\quad $^{\ddagger}$Massachusetts Institute of Technology\\
\ttt{\{tianyug,danqic\}@cs.princeton.edu}\\
\ttt{fisch@csail.mit.edu}
}

\date{}

\begin{document}
\maketitle
\renewcommand{\thefootnote}{\fnsymbol{footnote}}
\footnotetext[1]{The first two authors contributed equally.}
\renewcommand{\thefootnote}{\arabic{footnote}}

\input{sections/abstract}
\input{sections/intro}
\input{sections/related}

\input{sections/setup}
\input{sections/prompt}
\input{sections/context}
\input{sections/results}
%\input{sections/analysis}
\input{sections/limitations}
%\input{sections/discussion}
\input{sections/conclusion}

\clearpage
\bibliography{ref}
\bibliographystyle{acl_natbib}

\clearpage
\appendix
\counterwithin{figure}{section}
\counterwithin{table}{section}
\input{sections/appendix}

%\input{sections/appendix}

\end{document}

%% file: header.tex
\newcommand\ti[1]{\textit{#1}}

\newcommand\tf[1]{\textbf{#1}}
\newcommand\ttt[1]{\texttt{#1}}
\newcommand\mf[1]{\mathbf{#1}}

\newcommand{\ours}{LM-BFF\xspace}
\newcommand{\dtrain}{\mathcal{D}_{\text{train}}}
\newcommand{\ddev}{\mathcal{D}_{\text{dev}}}
\newcommand{\dtest}{\mathcal{D}_{\text{test}}}

\newcommand{\maskx}{\texttt{<X>}}
\newcommand{\masky}{\texttt{<Y>}}
\newcommand{\maskz}{\texttt{<Z>}}
\newcommand{\lsent}{\texttt{<}S_1\ttt{>}}
\newcommand{\lfirstsent}{\texttt{<}S_1\ttt{>}}
\newcommand{\lsecondsent}{\texttt{<}S_2\ttt{>}}
\newcommand{\sent}{\ttt{<}$S_1$\ttt{>}}
\newcommand{\firstsent}{\ttt{<}$S_1$\ttt{>}}
\newcommand{\secondsent}{\ttt{<}$S_2$\ttt{>}}
\newcommand{\xinput}{{x}_{\mathrm{in}}}
\newcommand{\xprompt}{x_{\mathrm{prompt}}}
\newcommand{\gen}{\mathrm{g}}

\newcommand{\template}{\mathcal{T}}

\newcommand{\lwordmap}{\mathcal{M}}
\newcommand{\vocabulary}{\mathcal{V}}
\newcommand{\labelset}{\mathcal{Y}}
\newcommand{\mapping}{\mathcal{M}}
\newcommand{\totalk}{K_{\text{tot}}}
\newcommand{\lm}{\mathcal{L}}
\newcommand{\cls}{\texttt{[CLS]}}
\newcommand{\sep}{\texttt{[SEP]}}
\newcommand{\mask}{\texttt{[MASK]}}

\renewcommand{\paragraph}[1]{\vspace{0.2cm}\noindent\textbf{#1}}
\newcommand{\tpf}[1]{\noindent\textbf{#1}}

\newcommand{\tableindent}{~~}

%% file: sections/abstract.tex
%!TEX root = ../main.tex

\begin{abstract}
The recent GPT-3 model~\cite{brown2020language} achieves remarkable few-shot performance solely by leveraging a natural-language prompt and a few task demonstrations as input context.
Inspired by their findings, we study few-shot learning in a more practical scenario, where we use smaller language models for which fine-tuning is computationally efficient.
We present {\ours}---\underline{b}etter \underline{f}ew-shot \underline{f}ine-tuning of \underline{l}anguage \underline{m}odels\footnote{Alternatively, {l}anguage {m}odels' \underline{b}est \underline{f}riends \underline{f}orever.}---a suite of simple and complementary techniques for fine-tuning language models on a small number of annotated examples. Our approach includes (1) prompt-based fine-tuning together with a novel pipeline for automating prompt generation; and (2) a refined strategy for dynamically and selectively incorporating demonstrations into each context.
Finally, we present a  systematic evaluation for analyzing few-shot performance on a range of NLP tasks, including classification and regression. Our experiments demonstrate that our methods combine to dramatically outperform standard fine-tuning procedures in this low resource setting, achieving up to 30\% absolute improvement, and 11\% on average across all tasks. Our approach makes minimal assumptions on task resources and domain expertise, and hence constitutes a strong task-agnostic method for few-shot learning.\footnote{Our implementation is publicly available at \url{https://github.com/princeton-nlp/LM-BFF}.}

\end{abstract}

%% file: sections/intro.tex
%!TEX root = ../main.tex

\section{Introduction}

\begin{figure*}[t]
    \centering
    \includegraphics[width=0.95\textwidth]{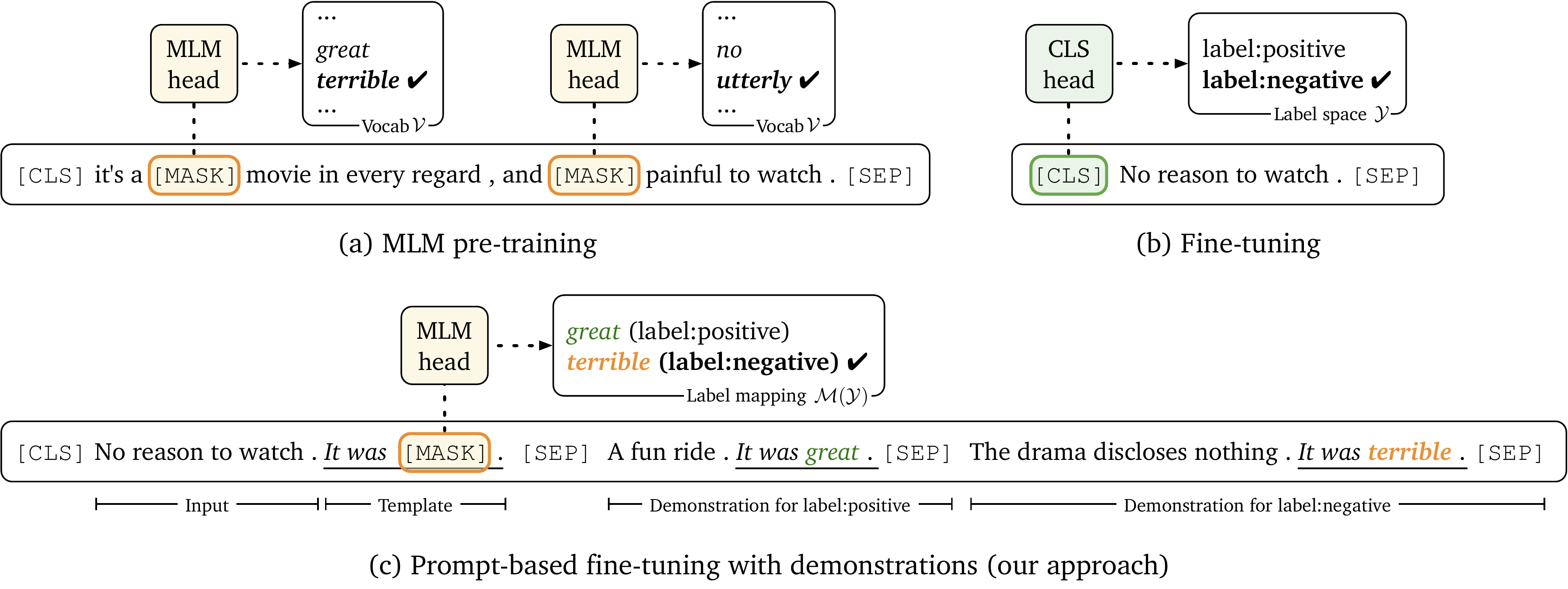}
    \caption{An illustration of (a) masked language model (MLM) pre-training, (b) standard fine-tuning, and (c) our proposed {\ours} using prompt-based fine-tuning with demonstrations. The underlined text is the task-specific \emph{template}, and colored words are \emph{label words}.}
    %\vspace{-5pt}
    \label{fig:overview}
\end{figure*}

\label{sec:intro}

The GPT-3 model \cite{brown2020language} has made waves in the NLP community by demonstrating astounding few-shot capabilities on myriad language understanding tasks.
Given only a \ti{natural language prompt} and a few \emph{demonstrations} of the task, GPT-3 is able to make accurate predictions without updating any of the weights of its underlying language model.
However, while remarkable, GPT-3 consists of 175B parameters, which makes it challenging to use in most real-wold applications.

In this work, we study a more practical scenario in which we only assume access to a moderately-sized language model such as BERT~\cite{devlin2019bert} or RoBERTa~\cite{liu2019roberta}, and a small number of examples (i.e., a \emph{few-shot} setting), which we can use to fine-tune the weights of
the language model.
This setting is appealing as
(1) such models can be trained on typical research hardware;
(2) few-shot settings are realistic, as it is generally both easy to acquire a few annotations (e.g., 32 examples) and efficient to train on them; and
(3)  updating parameters typically leads to better performance.
Inspired by GPT-3's findings, we propose several novel strategies for expanding its few-shot learning abilities to our setting, considering both classification and---for the first time---regression.

First, we follow the route of \ti{prompt-based} prediction, first developed by the GPT series~\cite{radford2018improving,radford2019language,brown2020language} for zero-shot prediction and recently studied by PET~\cite{schick2020exploiting, schick2020size} for fine-tuning.
Prompt-based prediction treats the downstream task as a (masked) language modeling problem, where the model directly generates a textual response (referred to as a \emph{label word}) to a given prompt defined by a task-specific \emph{template} (see Figure~\ref{fig:overview}(c)).
%\danqi{It is hard to see which part of Figure 1 refers to prompt-based fine-tuning.}
Finding the right prompts, however, is an art---requiring both domain expertise and an understanding of the language model's inner workings.
Even if significant effort is invested, manual prompts are likely to be suboptimal.
We address this issue by introducing automatic prompt generation, including a pruned brute-force search to identify the best working label words, and a novel decoding objective to automatically generate templates using the generative T5 model~\cite{raffel2020exploring}---all of which only require the few-shot training data. This allows us to cheaply obtain effective prompts that match or outperform our manually chosen ones.

Second, we adopt the idea of incorporating demonstrations %into each input
as
additional context.
GPT-3's naive ``in-context learning'' paradigm picks up to 32 randomly sampled examples, and concatenates them with the input. This method is not guaranteed to prioritize the most informative demonstrations, and mixing random examples from different classes together creates long contexts which can be hard to learn from. Additionally, the number of usable demonstrations is bounded by the model's maximum input length.
We develop a more refined strategy, where, for each input, we randomly sample a \emph{single} example at a time from \emph{each} class to create multiple, minimal demonstration \emph{sets}.
We also devise a novel sampling strategy that pairs inputs with similar examples, thereby providing the model with more discriminative comparisons.

We present a systematic evaluation for analyzing few-shot performance on 8 single-sentence and 7 sentence-pair NLP tasks. We observe that given a small number of training examples, (1) prompt-based fine-tuning largely outperforms standard fine-tuning; (2) our automatic prompt search method matches or outperforms manual prompts; and (3) incorporating demonstrations is effective for fine-tuning, and boosts few-shot performance.
Together, these simple-yet-effective methods contribute towards a dramatic improvement across the tasks we evaluate on,
and we obtain gains up to 30\% absolute improvement (11\% on average) compared to standard fine-tuning.
For instance, we find that a RoBERTa-large model achieves around 90\% accuracy on most binary sentence classification tasks, while only relying on 32 training examples. We refer to our approach as \ours, \underline{b}etter \underline{f}ew-shot \underline{f}ine-tuning of \underline{l}anguage \underline{m}odels: a strong, task-agnostic method for few-shot learning.

%% file: sections/related.tex
%!TEX root = ../main.tex

\section{Related Work}
\label{sec:related_work}

\tpf{Language model prompting.} The GPT series~\citep{radford2018improving,radford2019language,brown2020language} fueled the development of prompt-based learning,
and we follow many of its core concepts.
We are also greatly inspired by the recent PET work~\citep{schick2020exploiting,schick2020size}, although they mainly focus on a semi-supervised setting where a large set of unlabeled examples are provided.
We only use a few annotated examples as supervision, and also explore automatically generated prompts and fine-tuning with demonstrations.
Furthermore, we deviate from their evaluation by providing a more rigorous framework, as we will discuss in \S\ref{sec:setup}.
Finally, there is a large body of work on prompting for mining knowledge from pre-trained models \cite[][\emph{inter alia}]{trinh2018simple,petroni2019language,davison2019commonsense,talmor2020olmpics}. Different from these works, we focus on leveraging prompting for fine-tuning on downstream tasks.

\paragraph{Automatic prompt search.}
\citet{schick2020exploiting} and \citet{schick2020automatically} explore ways of identifying label words automatically,
however, none of these results lead to better performance compared to hand-picked ones.
In contrast, our method searches over both templates and label words, and is able to match or outperform our manual prompts.
Several other attempts have been made in addition---yet these approaches either operate in limited domains,
such as finding patterns to express specific relations~\cite{jiang2020can},
or require a large number of examples for gradient-guided search~\cite{shin2020autoprompt,zhong2021factual}. Our approach aims to develop general-purpose search methods that rely only on a few annotations.

%\paragraph{Fine-tuning of language models.} A number of recent studies have focused on better optimization and regularization tricks for stabilizing fine-tuning language models~\cite{howard2018universal,dodge2020fine,lee2020mixout, zhang2020revisiting}.
%Here we mainly use standard optimization techniques while focusing only
%on better prompt-based fine-tuning in a more extreme few-shot setting,
%and these studies are largely complementary to ours.

\paragraph{Fine-tuning of language models.} A number of recent studies have focused on better methods for fine-tuning language models~\cite{howard2018universal,dodge2020fine,lee2020mixout, zhang2020revisiting}. These works mainly focus on optimization and regularization techniques to stabilize fine-tuning. Here we use standard optimization techniques, and instead mainly focus our efforts on better prompt-based fine-tuning in a more extreme few-shot setting. We anticipate that results of these studies are largely complementary to ours.

\paragraph{Few-shot learning.} Broadly speaking, our setting is also connected to other few-shot learning paradigms in NLP, including
(1) {semi-supervised learning}~\cite{miyato2017adversarial,xie2020unsupervised,chen2020mixtext}, where a set of unlabeled examples are given;
(2) {meta-learning}~\cite{yu2018diverse,han2018fewrel,bansal2020learning,bansal2020self, bao2020fewshot}, where a set of auxiliary tasks are given; and
(3) {intermediate training}~\cite{phang2018sentence,yin2020universal}, where a related, intermediate task is given. We deviate from these settings by making minimal assumptions about available resources: we only assume a few annotated examples and a pre-trained language model. Our focus is on understanding how far we can push without any other advantages.

%% file: sections/setup.tex
%!TEX root = ../main.tex
%\input{tables/datasets}

\input{tables/manual_prompts}

\section{Problem Setup}
\label{sec:setup}

\paragraph{Task formulation.}
% \label{sec:setup_formulation}
In this work, we assume access to a pre-trained language model $\lm$ that we wish to fine-tune on a task $\mathcal{D}$ with a label space $\labelset$. For the task, we only assume $K$ training examples \emph{per class}\footnote{For regression, we partition the data into two ``classes'' according to being above or below the median value.} for the task's training set $\dtrain$, such that the total number of examples is $\totalk = K \times |\labelset|$, and $\dtrain = \{(\xinput^i, y^i)\}_{i=1}^{\totalk}$.
Our goal is then to develop task-agnostic learning strategies that generalize well to an unseen test set $(\xinput^{\text{test}}, y^{\text{test}})\sim \dtest$. %$(\xinput^{\text{test}}, y^{\text{test}}) \sim \dtest$.
For model selection and hyper-parameter tuning, we assume a development set $\ddev$, of the same size as the few-shot training set, i.e., $|\ddev| = |\dtrain|$. This distinction is important: using a larger development set confers a significant advantage
(see our experiments in Appendix~\ref{app:dev_size}),
and subverts our initial goal of learning from limited data.\footnote{In contrast, \newcite{schick2020exploiting,schick2020size} do not use a development set, and adopt a set of hyper-parameters based on practical considerations.
This is akin to ``shooting in the dark'' on a setting that we show can have unintuitive outcomes.}
For all of the following experiments (unless specified otherwise), we take $\lm=$ RoBERTa-large and $K=16$.

\paragraph{Evaluation datasets.}
% \label{sec:setup_dataset_and_protocol}
We conduct a systematic study across $8$ single-sentence and $7$ sentence-pair English tasks, %(see Table~\ref{tab:datasets}).
%Our evaluation is derived from
including
8 tasks from the GLUE benchmark~\cite{wang2019glue},
%\footnote{We follow previous work and exclude the WNLI task.}
SNLI~\cite{bowman2015large_snli}, and 6 other popular sentence classification tasks (SST-5, MR, CR, MPQA, Subj, TREC). All of the dataset details are provided in Appendix~\ref{app:datasets}. For \emph{single-sentence} tasks, the goal is to make a prediction based on an input sentence $\xinput = x_1$, such as whether a movie review is positive or not. For \emph{sentence-pair} tasks, the goal is to take a pair of input sentences $\xinput = (x_1, x_2)$ and predict the relationship between them. We also interchangeably refer to the inputs as {\firstsent} or (\firstsent, \secondsent).
% We refer to the single-sentence inputs as $\xinput=x_1$ or \firstsent, and the sentence-pair inputs as $\xinput=(x_1, x_2)$ or (\firstsent, \secondsent).
% We mainly use SST-2 and SNLI for pilot experiments and avoid tuning the model and prompts on other datasets, making it close to a true few-shot setting as much as possible.
Note that we mainly use SST-2 and SNLI for pilot experiments and model development, making it close to a true few-shot setting, at least for all the other datasets we evaluate on.

\paragraph{Evaluation protocol.}
Systematically evaluating few-shot performance can be tricky.
It is well-known that
fine-tuning on small datasets can suffer from instability~\cite{dodge2020fine,zhang2020revisiting}, and results may change dramatically given a new split of data.
% We find that fine-tuning results may change dramatically given different splits of few-shot data.
To account for this, we measure average performance across 5 different randomly sampled $\dtrain$ and $\ddev$ splits.
%using a fixed set of seeds $\seedset$.
This issue has also been discussed in
\newcite{schick2020size}---they suggest using a fixed set of training examples. We argue that sampling
%multiple $\dtrain$ and $\ddev$ gives a
multiple splits gives a
more robust measure of performance,
and a better estimate of the variance.
We also observe that hyper-parameters
%(as simple as the learning rate or batch size)
can make a significant difference,
%across model settings or data splits.
%thus for each set $\{\dtrain^s, \ddev^s\}, s\in \seedset$, we perform a grid search over multiple hyper-parameters, and take the best one as measured on $\ddev^s$.
thus we sweep multiple hyper-parameters for each data sample, and take the best setting as measured on the $\ddev$ of that sample (see Appendix~\ref{app:hyper_selection}).

%% file: tables/manual_prompts.tex
%!TEX root = ../main.tex

\begin{table*}[t]
\begin{center}
\centering
\resizebox{1.98\columnwidth}{!}{%
\begin{tabular}{lll}
\toprule
\tf{Task} & \tf{Template} & \tf{Label words}\\
\midrule
SST-2 &  {\sent} It was {\mask} . & positive: great, negative: terrible\\
SST-5 &  {\sent} It was {\mask} . & v.positive: great, positive: good, neutral: okay, negative: bad, v.negative: terrible\\
MR    & {\sent} It was {\mask} . & positive: great, negative: terrible\\
CR    & {\sent} It was {\mask} . & positive: great, negative: terrible\\
Subj  & {\sent} This is {\mask} . & subjective: subjective, objective: objective \\
TREC  & {\mask} : {\sent} & abbreviation: Expression, entity: Entity, description: Description \\
&& human: Human, location: Location, numeric: Number \\
COLA  & {\sent} This is {\mask} . & grammatical: correct, not\_grammatical: incorrect \\
\midrule
MNLI  & {\firstsent} ? {\mask} , {\secondsent} & entailment: Yes, netural: Maybe, contradiction: No \\
SNLI  & {\firstsent} ? {\mask} , {\secondsent} & entailment: Yes, netural: Maybe, contradiction: No\\
QNLI  & {\firstsent} ? {\mask} , {\secondsent} & entailment: Yes, not\_entailment: No \\
RTE   & {\firstsent} ? {\mask} , {\secondsent} & entailment: Yes, not\_entailment: No \\
MRPC  & {\firstsent} {\mask} , {\secondsent} & equivalent: Yes, not\_equivalent: No\\
QQP   & {\firstsent} {\mask} , {\secondsent} & equivalent: Yes, not\_equivalent: No\\
STS-B & {\firstsent} {\mask} , {\secondsent} & $y_u$: Yes, $y_l$: No \\
\bottomrule
\end{tabular}
}
\end{center}
\caption{Manual templates and label words that we used in our experiments. 
STS-B is a regression task (\S\ref{sec:regression}).
}
\label{tab:manual_prompts}
\vspace{-5pt}
\end{table*}

%% file: sections/prompt.tex
%!TEX root = ../main.tex

% \input{tables/manual_prompts}

% \section{Prompt-based Fine-tuning}
\section{Prompt-based Fine-tuning}
\label{sec:prompt_finetuning}

Given a masked language model
$\lm$,
we first convert input $\xinput$ to a token sequence $\tilde{x}$, and the language model $\lm$ then
%operates on $\tilde{x}$ and maps it to a sequence of hidden vectors $\mf{h}_k \in \mathbb{R}^d$.
maps $\tilde{x}$ to a sequence of hidden vectors $\{\mf{h}_k \in \mathbb{R}^d\}$.
During standard fine-tuning, we usually take $\tilde{x}_{\text{single}} = \cls x_1 \sep$ or $\tilde{x}_{\text{pair}} =  \cls x_1 \sep x_2 \sep$.
% Then, we can define $\lm(\tilde{x}_{\text{single}}) = (\mf{h}_{\cls}, \mf{h}_1, \mf{h}_2, \ldots, \mf{h}_{|x_1|}, \mf{h}_{\sep})$ (similarly for  $\lm(\tilde{x}_{\text{pair}})$).
For downstream classification tasks with a label space $\labelset$, we train a task-specific head, $\mathrm{softmax}(\mf{W}_o \mf{h}_{\cls})$, by maximizing the log-probability of the correct label, where $\mf{h}_{\cls}$ is the hidden vector of \cls, and $\mf{W}_o \in \mathbb{R}^{\mathcal{|\labelset|} \times d}$ is a set of randomly initialized parameters introduced at the start of fine-tuning.
%All of the parameters of $\lm$ as well as $\mf{W}_o$ are jointly fine-tuned to maximize the log-probability of the correct label.
Similarly, for a regression task, we can introduce $\mf{w}_o \in \mathbb{R}^d$ and optimize the mean squared error between $\mf{w}_o \cdot \mf{h}_{\cls}$ and the gold label.
In either case, the number of new parameters can be substantial---for example, a simple binary classification task will introduce 2,048 new parameters for a RoBERTa-large model---making it challenging to learn from a small amount of annotated data (e.g., 32 examples).

%$\mf{h}_{\sep}$, ($\mf{h}_{\underline{1}}$, $\mf{h}_{\underline{2}}$, $\ldots$, $\mf{h}_{\underline{|x_2|}}$, $\mf{h}_{\underline{\sep}}$) $\in \mathbb{R}^d$.

An alternative approach to solving this problem is \ti{prompt-based fine-tuning}, in which $\lm$ is directly tasked with ``auto-completing'' natural language prompts. % with the correct outputs.
For instance, we can formulate a binary sentiment classification task using a prompt with
%an input sentence $x_1$
input $x_1$
(e.g., ``\ti{No reason to watch it .}'') as:
% \begin{quoting}
%     $\xprompt$ = \cls $x_1$ {It was} \mask . \sep
% \end{quoting}
%\vspace{-5pt}
\begin{equation*}
    \resizebox{.85\hsize}{!}{%
    $\xprompt = \text{\cls~$x_1$~{It was}~\mask~. \sep}$
    }
\end{equation*}
and let $\lm$ decide whether it is more appropriate to fill in ``\emph{great}'' (positive) or ``\emph{terrible}'' (negative) for \mask.
We now formalize this approach for classification and regression (\S\ref{sec:classification} and \S\ref{sec:regression}), and discuss the importance of prompt selection (\S\ref{sec:manual_prompts}).

\subsection{Classification}
\label{sec:classification}

Let $\mapping \colon \labelset \rightarrow \vocabulary$ be a mapping from the task label space to individual words\footnote{More generally, we can consider a one-to-many mapping $\mapping\colon \labelset \rightarrow 2^{|\labelset|}$ in which we map labels to sets of words. However, we did not find significant gains in our experiments.}
% in our vocabulary $\vocabulary$, where $\vocabulary$ is given by the pre-trained language model $\lm$.
in the vocabulary $\vocabulary$ of $\lm$.
Then for each $\xinput$, let the manipulation ${x}_{\mathrm{prompt}} = \template(\xinput)$
%\footnote{Note that it is also possible to use auto-regressive language models, but we choose to focus on MLMs in this work.}
be a \emph{masked language modeling} (MLM) input which contains one \mask~token.
% where we have $\mathbf{x}_{\mathrm{prompt}}^{(m)} = \texttt{<mask>}$ at some index $m$.
In this way, we can treat our task as an
%a \emph{language modeling}
MLM, and model the probability of predicting class $y \in \labelset$ as:
\vspace{-10pt}
\begin{equation}
\label{eq:lm-classification}
\resizebox{.85\hsize}{!}{%
$\begin{aligned}
p(y \mid \xinput) &= p\left(\mask = \mapping(y) \mid \xprompt\right) \\
&=\frac{\exp\left(\mf{w}_{\mapping(y)} \cdot \mf{h}_{\mask}\right)}{\sum_{y' \in \labelset} {\exp\left(\mf{w}_{\mapping(y')} \cdot \mf{h}_{\mask}\right)}},
\end{aligned}$%
}
% \resizebox{.89\hsize}{!}{$\displaystyle
% p(y \mid \xinput) = \frac{\exp\left(\mf{w}_{\mapping(y)} \cdot \mf{h}_{\mask}\right)}{\sum_{y' \in \labelset} {\exp\left(\mf{w}_{\mapping(y')} \cdot \mf{h}_{\mask}\right)}},
% $}
\end{equation}
where $\mf{h}_{\mask}$ is the hidden vector of {\mask} %position obtained by applying $\lm$ on ${x}_{\mathrm{prompt}}$,
% and $\mf{w}_v$ denotes the pre-softmax output vector for any word used in pre-training, $v \in \vocabulary$.
and $\mf{w}_v$ denotes the pre-softmax vector corresponding to $v \in \vocabulary$.
When supervised examples $\{(\xinput, y)\}$ are available, $\mathcal{L}$ can be fine-tuned to minimize the cross-entropy loss.
It is important to note that this approach re-uses the pre-trained weights $\mf{w}_v$ and does not introduce any new parameters. It also reduces the gap between pre-training and fine-tuning, making it more effective in few-shot scenarios.

\subsection{Regression}
\label{sec:regression}
We assume the same basic setup as in classification, but treat the label space $\labelset$ as a bounded interval $[v_l, v_u]$.
%Inspired by recent work in prototypical few-shot regression~\cite{mettes2019hyperspherical},
%In regression, the label space $\labelset$ is a bounded interval $[v_l, v_u]$.
Inspired by ~\citet{mettes2019hyperspherical},
we model the problem as an interpolation between two opposing poles, $\{y_l, y_u\}$, with values $v_l$ and $v_u$ respectively.
For instance, we can formulate our previous sentiment analysis task as a regression problem in the range $[0, 1]$, where we slide between ``\emph{terrible}'' ($v_l = 0$) and ``\emph{great}'' ($v_u = 1$). In this way, we can express $y$ as a \emph{mixture model}:
\begin{equation}
    y = v_l \cdot p(y_l \mid \xinput) + v_u \cdot p(y_u \mid \xinput),
\end{equation}
where $p(y_u \mid \xinput)$ is the probability of $y_u$, and $p(y_l \mid \xinput) = 1 - p(y_u \mid \xinput)$.
%We define $\mapping$ as a map of $\{y_l, y_u\}$ to two words $\{w_l, w_u\} \subset \mathcal{V}$,
Then we define $\mapping \colon \{y_l, y_u\} \rightarrow \vocabulary$,
and model $p(y_u \mid \xinput)$ the same as Eq. (\ref{eq:lm-classification}).
\begin{comment}
\begin{equation}
\label{eq:lm-regression}
\resizebox{.89\hsize}{!}{$\displaystyle
p(y_u \mid \xinput) = \frac{\exp\left(\mf{w}_{w_u} \cdot \mf{h}_{\mask}\right)}{\sum_{w' \in \{w_u, w_l\}}\exp\left(\mf{w}_{w'} \cdot \mf{h}_{\mask}\right)}.
$}
\end{equation}
\end{comment}
%When supervised examples $\{(\xinput, y)\}$ are available,
We fine-tune $\mathcal{L}$ to minimize the KL-divergence between the inferred $p(y_u \mid \xinput)$
%of Eq.~\eqref{eq:lm-regression}
and the observed mixture weight, $(y
- v_l) / (v_u - v_l)$.

\input{tables/prompt_search}

\subsection{Manual prompts: the good and the bad}
\label{sec:manual_prompts}

%After formulating classification and regression tasks as MLM problems,
The key challenge is to construct the template $\template$ and label words $\mapping(\labelset)$---we refer to these two together as a \ti{prompt} $\mathcal{P}$.
% Considering the sentiment analysis task again, how can we tell that appending ``\ti{It was \mask .}'' after the input example $x_1$, and choosing between (\ti{great}, \ti{terrible}), will be a good choice for the task?
Previous works~\cite{schick2020exploiting,schick2020size} hand-craft both the templates and label words, which usually requires domain expertise and trial-and-error.
Table~\ref{tab:manual_prompts}
summarizes manual templates and label words chosen for each dataset in our experiments.
These templates and label words were designed by intuition, and by considering formats used in previous literature. %have been found to work well for similar tasks in the literature.

% Moreover, \newcite{schick2020exploiting} observed that even among a set of different, carefully chosen manual prompts, the best-performing one is often substantially better than the worst-performing one.

To better understand what constitutes a good template or label word, we conduct a pilot study on SST-2 and SNLI.
%Table~\ref{tab:prompt_search} shows that different prompts can have marked differences in final accuracy.
Table~\ref{tab:prompt_search} shows that different prompts can lead to substantial differences in final accuracy.
%Specifically, when a template is fixed, the better the label words match the ``semantic classes'',
Specifically, when a template is fixed, the better the label words match the ``semantic classes'',
the better the final accuracy is
(\ti{great}/\ti{terrible} $>$ \ti{good}/\ti{bad} $>$ \ti{cat}/\ti{dog}).
In extreme cases where we swap plausible label words (e.g., \ti{terrible}/\ti{great}), we achieve the worst overall performance.\footnote{It is unclear, however, why RoBERTa thinks that ``cat'' is more positive than ``dog''. The authors tend to disagree.}
% \danqi{It would be great if this footnote is not split into two columns :) }}
Furthermore, with the same set of label words,
even a small change in the template
%(e.g., removing punctuation)
can make a difference.
For example, for SNLI, if we put \mask~at the end, or swap sentence order, we observe a $>$10\% drop.
The above evidence clearly underlines the importance of selecting good templates and label words.
Searching for prompts, however, is hard, as the search space can be very large---especially for the template. Even worse, we only have a few examples to use to guide our search, which can easily overfit. We will address these issues next.

\section{Automatic Prompt Generation}
\label{sec:auto_prompt}

We now explore principled ways of automating the search process for label words (\S\ref{sec:label_search}) and templates (\S\ref{sec:template_search}). Our goals are to reduce the human involvement required to design prompts, and to find more optimal settings than those that we manually choose.
Here, we assume a classification task, but the process for regression is analogous.

\subsection{Automatic selection of label words}
\label{sec:label_search}
We first study how to construct a label word mapping $\mapping$ that maximizes accuracy on $\ddev$ after fine-tuning, given a fixed template $\template$.
Naively searching all possible assignments, however, is (1) generally intractable, as the search space is exponential in the number of classes; and (2) prone to overfitting, as we will tend to uncover spurious correlations given only a few annotations. As a simple solution,
for each class $c \in \labelset$, we construct a pruned set  $\mathcal{V}^c \subset \mathcal{V}$ of the top $k$ vocabulary words based on their conditional likelihood using the initial  $\lm$. That is, let $\dtrain^c \subset \dtrain$ be the subset of all examples of class $c$. We take $\mathcal{V}^c$ as
% \begin{equation}
% \resizebox{.89\hsize}{!}{$\displaystyle
% \mathcal{V}^c = \underset{v \in \mathcal{V}}{\mathrm{TopK}} \left\{\sum_{(\xinput, y) \in \dtrain^c} \hspace{-10pt}p(y \mid \xinput, \mapping(y) = v)\right\}
% $}
% \end{equation}
%\vspace{-5pt}
\begin{equation}
\resizebox{.89\hsize}{!}{$\displaystyle
 \underset{v \in \mathcal{V}}{\mathrm{Top}\text{-}k} \left\{\sum_{\xinput \in \dtrain^c} \hspace{-5pt}\log P_{\lm}\Big(\mask = v \mid \template(\xinput)\Big)\right\},
$}
\end{equation}
where ${P}_{\lm}$ denotes the output probability distribution of $\lm$.
To further narrow down the search space, we find the top $n$ assignments over the pruned space that maximize zero-shot accuracy on $\dtrain$ (both $n$ and $k$ are hyper-parameters, see
Appendix~\ref{app:prompts}).
Then we fine-tune all top $n$ assignments, and re-rank to find the best one using $\ddev$.
This approach is similar to the automatic verbalizer search methods in \newcite{schick2020exploiting,schick2020automatically}, except that we use a much simpler search process (brute-force) and also apply re-ranking---which we find to be quite helpful.

\subsection{Automatic generation of templates}
\label{sec:template_search}

Next, we study how to generate a diverse set of templates $\{\template\}$ automatically from a fixed set of label words $\mapping(\labelset)$.
To address this challenging problem, we propose to use T5~\cite{raffel2020exploring}, a large pre-trained text-to-text Transformer.
T5 is pre-trained to fill in missing spans (replaced by T5 mask tokens, e.g., \maskx~or \masky) in its input.
For example, given the input ``\ti{Thank you {\maskx} me to your party {\masky} week}'', T5 is trained to generate ``\ti{{\maskx} for inviting {\masky} last {\maskz}}'',
meaning that ``\ti{for inviting}'' is the replacement for {\maskx} and ``\ti{last}'' is the replacement for {\masky}.
This is well suited for prompt generation: we can simply take input sentences from $\dtrain$ and let the T5 model construct the template $\template$, without having to specify a pre-defined number of tokens for it.

Given an input example $(\xinput, y) \in \dtrain$, %, denote \maskx~and \masky~as T5 mask tokens, and $\lwordmap(y)$ as the label word for $y$.
we consider the following simple conversions, denoted as $\template_{\gen}(\xinput, y)$, for formulating the T5 model inputs:\footnote{We consider putting the label word both before and after the input sentence for single-sentence tasks. However, we find that it is always better to put the label words in the middle (between the two sentences) for sentence-pair tasks.}
%\vspace{-10pt}
\begin{equation*}
\resizebox{.85\hsize}{!}{%
$\begin{aligned}
\lsent &\longrightarrow~\maskx~\lwordmap(y)~\masky~\lsent, \\
\lsent &\longrightarrow~\lsent~\maskx~\lwordmap(y)~\masky,\\
\lfirstsent,\lsecondsent &\longrightarrow \lfirstsent~\maskx~\lwordmap(y)~\masky~\lsecondsent.
\end{aligned}$%
}
% \resizebox{.89\hsize}{!}{$\displaystyle
% p(y \mid \xinput) = \frac{\exp\left(\mf{w}_{\mapping(y)} \cdot \mf{h}_{\mask}\right)}{\sum_{y' \in \labelset} {\exp\left(\mf{w}_{\mapping(y')} \cdot \mf{h}_{\mask}\right)}},
% $}
\end{equation*}

\begin{figure}[t]
    \centering
    \includegraphics[width=0.48\textwidth]{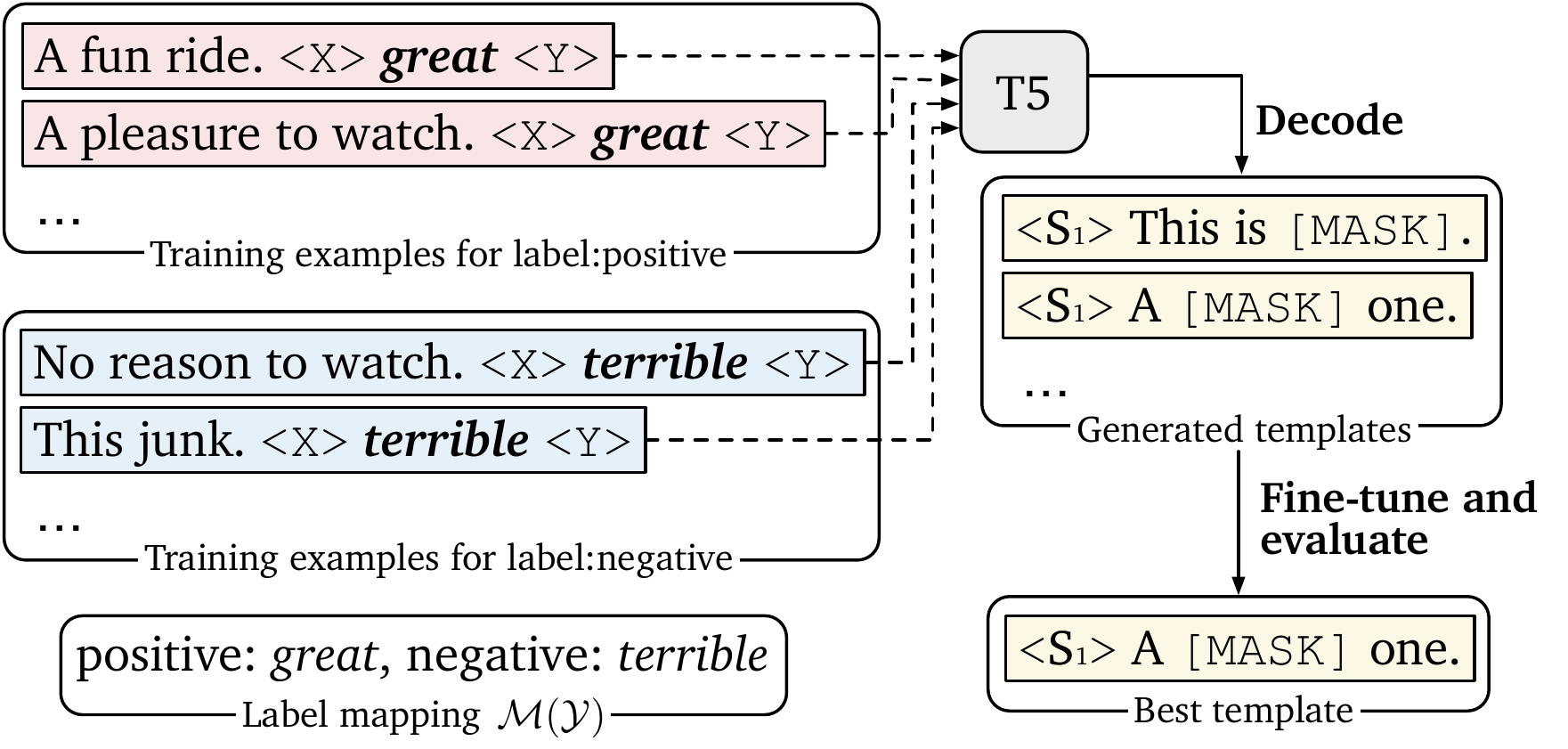}
    \caption{Our approach for template generation.
    }
    \label{fig:template_search}
    \vspace{-5pt}
    % there is also a template_search6.pdf showing label word in a different way
\end{figure}

As shown in Figure~\ref{fig:template_search}, we rely on the T5 model to fill in the placeholders. %, \maskx~and \masky.
% For instance, it might generate ``\ti{{\maskx} It was {\masky} .}'' for single-sentence tasks, or ``\ti{{\maskx} ? {\masky} , {\maskz}}'' for sentence-pair tasks.
When decoding, our goal here is to find an output that can work well for \emph{all} examples in $\dtrain$,
i.e.,
%Formally, we want
the output template $\template$ that maximizes $\sum_{(\xinput, y) \in \dtrain}{\log P_{\text{T5}}(\template \mid \template_{\gen}(\xinput, y))}$,
where $P_{\text{T5}}$ denotes the output probability distribution of T5.
It can be decomposed according to:
\vspace{-0.5em}
\begin{equation}
\resizebox{.87\hsize}{!}{$\displaystyle
    \sum_{j = 1}^{|\template|}\hspace{-10pt}\sum_{~~~~~(\xinput, y) \in \dtrain} {\hspace{-14pt}\log{P_{\text{T5}}\big(t_j \mid t_1,...,t_{j-1}, \template_{\gen}\big(\xinput,y\big)\big)}},
$}
\end{equation}
where $(t_1, \ldots, t_{|\template|})$ are the template tokens.

We use beam search to decode multiple template candidates. Concretely, we use a wide beam width (e.g., 100) to cheaply obtain a large set of diverse templates. We then fine-tune each generated template on $\dtrain$ and use $\ddev$ to either pick the single template with the best performance (Table~\ref{tab:main_results}), or  the top $k$ templates to use as an ensemble (Table~\ref{tab:ensemble}). Though it might appear to be expensive to fine-tune the model on each individual template, this is fast in practice due to the small size of $\dtrain$, and is also fully automated: making it easy to use, compared to manually tuning prompts for each dataset.

%% file: tables/prompt_search.tex
%!TEX root = ../main.tex

% Test results

\begin{table}[!t]
    \centering
    \resizebox{0.95\columnwidth}{!}{%
    \begin{tabular}{l cc c}
        \toprule
        \tf{Template} & \tf{Label words} & \tf{Accuracy} \\
        \midrule
        \multicolumn{2}{l}{SST-2 (positive/negative)} & mean (std)\\
        \midrule
        {\sent} It was {\mask} . & great/terrible & \tf{92.7 (0.9)} \\
        {\sent} It was {\mask} . & good/bad & 92.5 (1.0) \\
        {\sent} It was {\mask} . & cat/dog & 91.5 (1.4) \\
        {\sent} It was {\mask} . & dog/cat & 86.2 (5.4) \\
        {\sent} It was {\mask} . & terrible/great & 83.2 (6.9) \\
        {Fine-tuning} & - & 81.4 (3.8) \\
        \midrule
        \multicolumn{2}{l}{SNLI (entailment/neutral/contradiction)} & mean (std)\\
        \midrule
        {\firstsent} ? {\mask} , {\secondsent} & Yes/Maybe/No & \tf{77.2 (3.7)} \\
        {\firstsent} . {\mask} , {\secondsent} & Yes/Maybe/No & 76.2 (3.3) \\
        {\firstsent} ? {\mask} {\secondsent} & Yes/Maybe/No & 74.9 (3.0) \\
        {\firstsent} {\secondsent} {\mask} & Yes/Maybe/No &  65.8 (2.4) \\
        {\secondsent} ? {\mask} , {\firstsent} & Yes/Maybe/No & 62.9 (4.1) \\
        {\firstsent} ? {\mask} , {\secondsent} & Maybe/No/Yes & 60.6 (4.8) \\
        {Fine-tuning} & - & 48.4 (4.8) \\
        \bottomrule
    \end{tabular}
    }
    \caption{The impact of templates and label words on prompt-based fine-tuning ($K=16$).
    }
    \label{tab:prompt_search}
\end{table}

%% file: sections/context.tex
%!TEX root = ../main.tex

\section{Fine-tuning with Demonstrations}
\label{sec:demonstrations}

\input{tables/main_results}

In this section, we study whether we can leverage demonstrations when \ti{fine-tuning} medium-sized LMs, and find better ways to exploit them.

\subsection{Training examples as demonstrations}
GPT-3's naive approach to in-context learning simply involves concatenating the input with up to 32 examples randomly drawn from the training set.
This approach is suboptimal as
(1) the number of available demonstrations is bounded by the model's maximum input length;\footnote{GPT-3 uses a context size of 2,048 while most smaller language models (e.g., RoBERTa) have a context size of 512.}
%and hence including uninformative demonstrations is wasteful;
and (2) mixing numerous random examples from different classes together creates extremely long contexts which can be hard to leverage, especially for a smaller model.
To address these issues,
we propose a simpler solution:
at each training step,
we randomly sample \ti{one}\footnote{We also explored sampling multiple examples per class, but did not observe any improvements.} example $\big(\xinput^{(c)}, y_{\phantom{t}}^{(c)}\big)\in \dtrain$ from each class,
convert it into $\template\big(\xinput^{(c)}\big)$ with {\mask} replaced by $\mapping(y_{\phantom{t}}^{(c)})$---we denote this as $\tilde{\template}\big(\xinput^{(c)}, y_{\phantom{t}}^{(c)}\big)$---and then concatenate them with $\xinput$ (Figure~\ref{fig:overview}(c)):
%\vspace{-5pt}
\begin{equation*}
\resizebox{.89\hsize}{!}{$\displaystyle
    \template\big(\xinput\big) \oplus \tilde{\template}\big(\xinput^{(1)}, y_{\phantom{t}}^{(1)}\big) \oplus \cdots \oplus \tilde{\template}\big(\xinput^{(|\labelset|)}, y_{\phantom{t}}^{(|\labelset|)}\big).
$}
%\vspace{-3pt}
\end{equation*}
Here $\oplus$ denotes concatenation of input sequences.
During both training and inference we sample multiple demonstration sets for each $\xinput$. Note that both $\xinput$ and demonstration examples are sampled from the same set $\dtrain$ during training. At testing time, we still sample demonstration sets from $\dtrain$ and ensemble predictions across all sets.

% We update on each instance independently during training, while we ensemble predictions across all sets at inference time.

\subsection{Sampling similar demonstrations}
\label{sec:demonstration_sampling}
We observe that controlling the construction of the demonstration examples $\{(\xinput^{(c)}, y_{\phantom{t}}^{(c)})\}$ is crucial for good final performance.
For example, if the set of contrastive demonstrations $\xinput^{(c)}$ are all dramatically different---from each other, or from the query $\xinput$---then it becomes challenging for the language model to decipher meaningful patterns. As a result, the model may simply ignore the context, or even get confused by the additional examples.
To address this issue, we devise a simple strategy in which we only sample examples that are semantically close to $\xinput$.
Specifically, we use a pre-trained SBERT~\cite{reimers2019sentence} model to obtain embeddings for all input sentences (for sentence-pair tasks, we use the concatenation of the two sentences).
Here we just feed the raw sentences without the templates into SBERT.
For each query $\xinput$ and each label $c\in \labelset$, we sort all training instances with the label $x \in \dtrain^{c}$ by their similarity score to the query $\cos(\mf{e}(\xinput), \mf{e}(x))$, and only sample from the top $r = 50\%$ instances for each class to use as demonstrations.

%% file: tables/main_results.tex
%!TEX root = ../main.tex

% MRPC/QQP with f1
\begin{table*}[t]
\begin{center}
\centering
\resizebox{1.0\textwidth}{!}{%
\begin{tabular}{lcccccccc}
\toprule
& \tf{SST-2} & \tf{SST-5} & \tf{MR} & \tf{CR} & \tf{MPQA} & \tf{Subj} &  \tf{TREC} & \tf{CoLA} \\
& (acc) & (acc) & (acc) & (acc) & (acc) & (acc) & (acc) & (Matt.)\\
\midrule
% single sentence tasks
Majority$^\dagger$ & \ti{50.9} & \ti{23.1} & \ti{50.0} & \ti{50.0} & \ti{50.0} & \ti{50.0} & \ti{18.8} & \ti{0.0}  \\
Prompt-based zero-shot$^\ddagger$ & 83.6  &	35.0  &	80.8 &	79.5 &	67.6  &	51.4 &	32.0  &	2.0  \\
% \midrule
``GPT-3'' in-context learning &84.8 (1.3) &	30.6 (0.9) &	80.5 (1.7) &	87.4 (0.8) &	63.8 (2.1) &	53.6 (1.0) &	26.2 (2.4) &	-1.5 (2.4) \\
Fine-tuning & 81.4 (3.8) &	43.9 (2.0) &	76.9 (5.9) &	75.8 (3.2) &	72.0 (3.8) &	90.8 (1.8) &	{88.8} (2.1) &	\tf{33.9} (14.3) \\

\midrule
Prompt-based FT (man) & 92.7 (0.9) &	47.4 (2.5) &	87.0 (1.2) &	90.3 (1.0) &	84.7 (2.2) &	91.2 (1.1) &	84.8 (5.1) &	9.3 (7.3) \\
\tableindent + demonstrations & 92.6 (0.5) &	\tf{50.6} (1.4) &	86.6 (2.2) &	90.2 (1.2) &	\tf{87.0} (1.1) &	\tf{92.3} (0.8) &	87.5 (3.2) &	18.7 (8.8) 	\\
Prompt-based FT (auto) & 92.3 (1.0) &	49.2 (1.6) &	85.5 (2.8) &	89.0 (1.4) &	85.8 (1.9) &	91.2 (1.1) &	88.2 (2.0) &	14.0 (14.1) \\
\tableindent + demonstrations & \tf{93.0} (0.6) &	49.5 (1.7) &	\tf{87.7} (1.4) &	\tf{91.0} (0.9) &	86.5 (2.6) &	91.4 (1.8) &	\tf{89.4} (1.7) &	21.8 (15.9)\\
\midrule
Fine-tuning (full)$^\dagger$ & \ti{95.0} & \ti{58.7} & \ti{90.8} & \ti{89.4} & \ti{87.8} & \ti{97.0} & \ti{97.4} & \ti{62.6} \\

\midrule
    & \tf{MNLI} & \tf{MNLI-mm}  & \tf{SNLI} & \tf{QNLI} &  \tf{RTE} & \tf{MRPC} & \tf{QQP} & \tf{STS-B} \\
    & (acc) & (acc) & (acc) & (acc) & (acc) & (F1) & (F1) & (Pear.)\\
\midrule
% sentence pair tasks
Majority$^\dagger$ & \ti{32.7} & \ti{33.0} & \ti{33.8} & \ti{49.5} & \ti{52.7} & \ti{81.2}  & \ti{0.0} & \ti{-}  \\
Prompt-based zero-shot$^\ddagger$ &	50.8  &	51.7 &	49.5  &	50.8  &	51.3  & 61.9  &	49.7  &	-3.2   \\
% \midrule
``GPT-3'' in-context learning  & 52.0 (0.7) &	53.4 (0.6) &	47.1 (0.6) &	53.8 (0.4) &	60.4 (1.4) &	45.7 (6.0) &	36.1 (5.2) &	14.3 (2.8) \\
Fine-tuning  &	45.8 (6.4) &	47.8 (6.8) &	48.4 (4.8) &	60.2 (6.5) &	54.4 (3.9) & {76.6} (2.5) &	60.7 (4.3) &	53.5 (8.5) \\

\midrule
Prompt-based FT (man) &	68.3 (2.3) &	70.5 (1.9) &	77.2 (3.7) &	64.5 (4.2) &	69.1 (3.6) & 74.5 (5.3) &	65.5 (5.3) &	71.0 (7.0)  \\
\tableindent + demonstrations &	\tf{70.7} (1.3) &	\tf{72.0} (1.2) &	\tf{79.7} (1.5) &	\tf{69.2} (1.9) &	68.7 (2.3) & 77.8 (2.0) &	\tf{69.8} (1.8) &	73.5 (5.1)    	\\
Prompt-based FT (auto)  &	68.3 (2.5) &	70.1 (2.6) &	77.1 (2.1) &	68.3 (7.4) &	\tf{73.9} (2.2) & 76.2 (2.3) &	67.0 (3.0) &	75.0 (3.3) \\
\tableindent + demonstrations & 70.0 (3.6) &	\tf{72.0} (3.1) &	77.5 (3.5) &	68.5 (5.4) &	{71.1} (5.3) &	\tf{78.1} (3.4) &	67.7 (5.8) &	\tf{76.4} (6.2)  \\
\midrule
Fine-tuning (full)$^\dagger$ & \ti{89.8} &	\ti{89.5} &  \ti{92.6}	 &	\ti{93.3} &	\ti{80.9} &	\ti{91.4} &	\ti{81.7} & \ti{91.9} \\

\bottomrule
\end{tabular}}
\end{center}

\caption{
Our main results using RoBERTa-large.
$\dagger$: full training set is used (see dataset sizes in
Table~\ref{tab:datasets});
$\ddagger$: no training examples are used; otherwise we use
% $K = 16$ (\# examples per class)
$K = 16$ (per class)
for few-shot experiments.
% and $K=16$ (\# of training examples per class).
We report mean (and standard deviation) performance over 5 different splits (\S \ref{sec:setup}).
{Majority:} majority class;
{FT:} fine-tuning;
{man:} manual prompt
(Table~\ref{tab:manual_prompts});
{auto:} automatically searched templates (\S\ref{sec:template_search});
{``GPT-3'' in-context learning:} using the in-context learning proposed in \newcite{brown2020language} with RoBERTa-large (no parameter updates).
}
\label{tab:main_results}
\end{table*}

%% file: sections/results.tex
%!TEX root = ../main.tex

\section{Experiments}
\label{sec:experiments}
We present our main results, and address several research questions pertaining to our \ours approach.
Implementation details are in Appendix~\ref{app:exp_details}.

\subsection{Main results}
\label{sec:mainresult}
We use a RoBERTa-large model and set $K = 16$ in our experiments.
A comparison of using RoBERTa vs BERT can be found in
Appendix~\ref{app:analysis_bert}.
For automatic prompt search, in our main table we report automatic template search only (which consistently performs the best, see Table~\ref{tab:auto_search}). To put our results in perspective, we compare to a number of baselines, namely
(1) standard fine-tuning in our few-shot setting;
(2) standard fine-tuning using the full training set;
(3) simply taking the most frequent class (measured on the full training set);
(4) prompt-based zero-shot prediction where we take our manual prompts and use $\lm$ ``out-of-the-box'' without using any training examples; and
(5) ``GPT-3'' in-context learning, where we use the same prompt-based zero-shot setting, but augment the context with randomly sampled 32 demonstrations
(and still use RoBERTa-large, not GPT-3).

\input{tables/ensemble.tex}

\input{tables/auto_search_compare.tex}

\paragraph{Single-prompt results.}
Table~\ref{tab:main_results} shows our main results using a single prompt, either from our manually designed ones (Table~\ref{tab:manual_prompts}) , or the best generated ones.
First, prompt-based zero-shot prediction achieves much better performance than the majority class, showing the pre-encoded knowledge in RoBERTa.
Also, ``GPT-3'' in-context learning
does not always improve over zero-shot prediction, likely because smaller language models are not expressive enough to use off-the-shelf like GPT-3. % few-shot learning.

Second, prompt-based fine-tuning can greatly outperform standard fine-tuning, both when using a manual prompt or a generated one.
% CoLA (the linguistic acceptability task) is one exception, which is an interesting case as the input may be a non-grammatical sentence, and never seen during pre-training.
CoLA is one interesting exception, as the input may be a non-grammatical sentence which is out of the distribution of $\lm$. %, and never seen during pre-training.
Generally, our automatically searched templates can achieve comparable or even higher results than manual ones,
especially for tasks in which constructing strong manual templates is less intuitive (e.g., TREC, QNLI and MRPC).

\input{tables/generated_template.tex}

Finally, using demonstrations in context leads to consistent gains in a majority of tasks.
%Third, using demonstrations further improve the performance. %leads to consistent gains in a majority of tasks.
% they demonstrate that using basic manual prompt-based fine-tuning greatly outperforms our other baselines, including standard fine-tuning, and static approaches that do not update the underlying model (prompt-based zero-shot and GPT-3-style in-context learning).
 % Second, we see that using our automatically searched templates can achieve comparable or even higher results than manual templates.  Finally, in-context fine-tuning further boosts performance.
%It shows that in fine-tuning, providing demonstrations in the context also helps the model, by allowing it to compare the input with the example from each class.
In summary, our combined solution---fine-tuning with automatically searched templates and sampled demonstration sets---achieves a $30\%$ gain on SNLI compared to standard fine-tuning, and $11\%$ gain on average.
% over all datasets.

\paragraph{Ensemble results.} An advantage of automatic prompt search is that we can generate as many prompts as we want, train individual models, and create large ensembles. PET~\cite{schick2020exploiting,schick2020size} also ensembles multiple models trained with manual prompts.\footnote{They then use unlabeled data and distillation to get a single model, which is outside of our scope.} In Table~\ref{tab:ensemble}, we make a direct comparison of our searched prompts and PET's manual prompts on MNLI and RTE (two datasets that we evaluate in common).\footnote{In the PET NLI templates, the hypothesis is put before the premise, which we actually found to be suboptimal. In our experiments, we swap the two and get better results.}
As the results show, an ensemble with multiple templates always improves performance.
An ensemble of the same number of automatic templates achieves comparable or better performance than the ensemble of PET's manual prompts. Increasing the number of automatic templates brings further gains.

\subsection{Analysis of generated prompts}
Table~\ref{tab:auto_search} gives the results of using manual vs automatic prompts.
For automatic prompts, we compare
% template search only using manual label words (Auto T),
% label word search only using manual templates (Auto L),
template search (Auto T),
label word search (Auto L),
and a joint variant (Auto T + L) in which %we search both templates and label words one after the other (
we start from manual label words, apply Auto T, and then Auto L.
%and a joint variant (Auto T + L) in which we search both templates and label words one after the other (starting from a fixed set of manual label words, we apply Auto T, and then Auto L).
%Auto L outperforms manual prompts on TREC, RTE and MRPC---but is considerably worse on SNLI.
%As shown, though auto L performs better than manual ones sometimes, this trend is not consistent, and auto T + L is often better than auto L.
%As shown, auto L, though in most cases peforms comparably to the manual label words, can suffere a drop on certain tasks.
%In most cases, auto T achieves comparable or higher performance than manual prompts, and is the best variant overall.
%Our main automatic search results in Table~\ref{tab:main_results} use Auto T.\danqi{Is it too late to put here?}
%Table~\ref{tab:generated_template} shows several outputs from Auto T and Auto L (A full list can be found in Appendix~\ref{app:generated_prompts}).
In most cases, Auto T achieves comparable or higher performance than manual ones, and is consistently the best variant. Auto L outperforms manual prompts on TREC and MRPC---but is considerably worse on SNLI. Auto T + L is often better than Auto L, but only sometimes better than Auto T.
%We also list generated examples for auto T and auto L in
%Appendix~\ref{app:generated_prompts}.
%Appendix~\ref{app:generated_prompts}.
Table~\ref{tab:generated_template} shows examples from Auto T and Auto L (A full list in Appendix~\ref{app:generated_prompts}).
Auto T templates generally fit the context and label words well, but can contain biased peculiarities (e.g., ``\emph{\{Yes/No\}, no}'' in SNLI).
For Auto L words, things are mixed: while most look intuitively reasonable, there are also some mysterious abnormalities (e.g., ``\emph{Hi}'' for the ``entailment'' class in SNLI).

\subsection{Analysis of demonstration sampling}
\input{tables/pair_ablation}

Table~\ref{tab:pair_ablation} compares the performance of demonstrations using uniform sampling to selective sampling by SBERT.
%, across a number of selected tasks.
We acknowledge that SBERT is trained on SNLI and MNLI datasets, thus we also tried a simple sentence encoder using mean pooling of hidden representations from RoBERTa-large.
% We find that on SST-2 and SNLI (where prompt-based FT performs well),
% using selective sampling outperforms the uniform one, highlighting the importance of sampling similar examples for demonstrations.
We find that in either case, using selective sampling outperforms uniform sampling, highlighting the importance of sampling similar examples for incorporating demonstrations in context.
% \citet{liu2021makes} also took a selective sampling approach of GPT-3's in-context learning and come to a similar observation as ours.

\subsection{Sample efficiency}

%Taking a step back, it is interesting to see how standard fine-tuning and our method (\ours) compare as $K$ increases.
%Figure~\ref{fig:sample} illustrates the trends for SST-2 and SNLI.
Figure~\ref{fig:sample} illustrates
how standard fine-tuning and our \ours compare as $K$ increases.
For a simple task such as SST-2 (also see MR, CR and MPQA in Table~\ref{tab:main_results}), despite using only 32 total examples, \ours has already nearly saturated its performance and is comparable to standard fine-tuning over the entire dataset. On the harder task of SNLI, \ours continues to improve as $K$ increases while still maintaining a performance gap over standard fine-tuning, until the two converge around $K = 256$.

%% file: tables/ensemble.tex
%!TEX root = ../main.tex

\begin{table}[t]
    \begin{center}
    \centering
    \resizebox{0.93\columnwidth}{!}{%
    \begin{tabular}{lcc}
    \toprule
     {Prompt-based Fine-tuning}   & \tf{MNLI} & \tf{RTE} \\
    \midrule
    Our single manual $\mathcal{P}$ & 68.3 (2.3) & 69.1 (3.6) \\
    $\mathcal{P}_{\text{PET}}$ &  71.9 (1.5)  & 69.2 (4.0) \\
    $\mathcal{P}_{\text{ours}}$, $|\mathcal{P}_{\text{ours}}| = |\mathcal{P}_{\text{PET}}|$ & 70.4 (3.1) &	73.0 (3.2)\\
    % In-context FT (raw) (auto T ens) & \tf{73.6} (1.8) & \tf{75.3} (1.6) &  \tf{73.6} (3.4) \\
    \tableindent + demonstrations & 74.0 (1.9) &  71.9 (4.6)\\
    $\mathcal{P}_{\text{ours}}$, $|\mathcal{P}_{\text{ours}}| = 20$&72.7 (2.5) & \tf{73.1} (3.3)\\
    \tableindent + demonstrations& \tf{75.4} (1.6) & 72.3 (4.5)\\
    \bottomrule
    \end{tabular}
    }
    \end{center}
    \caption{Ensemble models using manual prompts from PET~\cite{schick2020exploiting,schick2020size} and our automatic templates. PET uses 4 prompts for MNLI and 5 for RTE. We also use an equal number of templates  
    in $|\mathcal{P}_{\text{ours}}| = |\mathcal{P}_{\text{PET}}|$ 
    for a fair comparison.}
    \vspace{-3pt}
    \label{tab:ensemble}
\end{table}

%% file: tables/auto_search_compare.tex
% Test results
%!TEX root = ../main.tex

\begin{table}[t]
    \begin{center}
    \centering
    \resizebox{0.90\columnwidth}{!}{%
    \begin{tabular}{lcccc}
    \toprule
             & \tf{SST-2}  & \tf{SNLI} & \tf{TREC}    & \tf{MRPC} \\
    \midrule
    Manual   & \tf{92.7}    & \tf{77.2}  & 84.8&74.5    \\
    \midrule
    Auto T   & 92.3    & 77.1  & {88.2} &76.2  \\
    Auto L   & 91.5     & 75.6 & 87.0 & \tf{77.2} \\
    Auto T + L & 92.1      & 77.0   & \tf{89.2} &74.0 \\
    \bottomrule
    \end{tabular}
    }
    \end{center}

    \caption{
        Comparison between manual prompts and different automatic prompt generation methods: auto-generated templates (Auto T), auto-generated label words (Auto L), and their combination (Auto T + L).
        % \tianyu{Do we want to keep T+L?}
        % \danqi{On a second thought, we can keep it....}
        % \tianyu{I removed T+L here since we didn't discuss this in the method part.}
        }

    \vspace{-3pt}
    \label{tab:auto_search}
\end{table}

% \begin{table}[t]
%     \begin{center}
%     \centering
%     \resizebox{0.90\columnwidth}{!}{%
%     \begin{tabular}{lcccc}
%     \toprule
%              & \tf{SST-2}  & \tf{TREC}   & \tf{SNLI}  & \tf{MRPC} \\
%     \midrule
%     Manual   & \tf{92.7}  & 84.8  & \tf{77.2}  &74.5    \\
%     \midrule
%     Auto T   & 92.3   & {88.2}  & 77.1  &76.2  \\
%     Auto L   & 91.5  & 87.0   & 75.6  & \tf{77.2} \\
%     Auto T + L & 92.1   & \tf{89.2}   & 77.0  &74.0 \\
%     \bottomrule
%     \end{tabular}
%     }
%     \end{center}
% 
%     \caption{
%         Comparison between manual prompts and different automatic prompt generation methods: auto-generated templates (Auto T), auto-generated label words (Auto L), and their combination (Auto T + L).
%         % \tianyu{Do we want to keep T+L?}
%         % \danqi{On a second thought, we can keep it....}
%         % \tianyu{I removed T+L here since we didn't discuss this in the method part.}
%         }
% 
% 
%     \label{tab:auto_search}
% \end{table}
% 

%% file: tables/generated_template.tex
%!TEX root = ../main.tex

\begin{table}[!t]
    \begin{center}
    \centering
    \resizebox{0.92\columnwidth}{!}{%
    \begin{tabular}{ll}
    \toprule
    \tf{SST-2} & (positive/negative)  \\
    \midrule

    Auto T & $\mapping(\labelset)$  = \{great, terrible\}\\
    & \#1. {\sent} A {\mask} one .\\
    & \#2. {\sent} A {\mask} piece .\\
    & \#3. {\sent} All in all {\mask} .\\
    \midrule
    Auto L & $\template(\xinput)$ = \sent~It was \mask . \\
    & \#1. irresistible/pathetic\\
    & \#2. wonderful/bad \\
    & \#3. delicious/bad \\

%     \midrule
%     \tf{MRPC} & (equivalent/not\_equivalent) \\
%     \midrule
%     Auto T & $\mapping(\labelset)$  = \{Yes, No\} \\
%      & \#1. {\firstsent} . {\mask} ! {\secondsent}  \\
%      & \#2. {\firstsent} . {\mask} . This is the first time {\secondsent}  \\
%      & \#3. {\firstsent} . {\mask} . That's right . {\secondsent}  \\
%     \midrule
%     Auto L & $\template(\xinput)$ = \firstsent \mask , \secondsent \\
%      & \#1. Rather/Alas \\
%      & \#2. At/Thus \\
%      & \#3. Instead/Moreover \\

    \midrule
     \tf{SNLI} & (entailment/neutral/contradiction) \\
     \midrule
    Auto T & $\mapping(\labelset)$  = \{Yes, Maybe, No\} \\
     & \#1. {\firstsent} . {\mask} , no , {\secondsent} \\
     & \#2. {\firstsent} . {\mask} , in this case {\secondsent} \\
     & \#3. {\firstsent} . {\mask} this time {\secondsent}  \\
    \midrule
    Auto L & $\template(\xinput)$ = \firstsent~? \mask~, \secondsent \\
     & \#1. Alright/Watch/Except \\
     & \#2. Hi/Watch/Worse \\
     & \#3. Regardless/Fortunately/Unless  \\
    \bottomrule
    \end{tabular}
    }
    \end{center}
    \vspace{-3pt}
    \caption{Examples of our automatically generated templates (Auto T) and label words (Auto L).}
    \vspace{-8pt}
    \label{tab:generated_template}
    \end{table}

%% file: tables/pair_ablation.tex
%!TEX root = ../main.tex

% Test, MRPC=f1
\begin{table}[t]
    \begin{center}
    \centering
    \resizebox{1.0\columnwidth}{!}{%
    \begin{tabular}{lcccc}
    \toprule
                       & \tf{SST-2} &   \tf{SNLI} &\tf{TREC} &  \tf{MRPC} \\
    \midrule
    Prompt-based FT    & \tf{92.7}        & 77.2  & 84.8     & 74.5   \\
    \midrule
    Uniform sampling     & 92.3            & 78.8  & 85.6     & 70.9 \\
    \tableindent + RoBERTa sel. & \tf{92.7}        & 79.5  & 83.4    & 76.6           \\
    \tableindent + SBERT sel.   & 92.6        & \tf{79.7}  & \tf{87.5} & \tf{77.8}   \\
    \bottomrule
    \end{tabular}
    }
    \end{center}

    \caption{
        Impact of demonstration sampling strategies.
        Uniform sampling randomly samples demonstrations, while selective (sel.) sampling only takes top sentences measured by the sentence encoders (\S \ref{sec:demonstrations}).
    }
    \vspace{-10pt}
    \label{tab:pair_ablation}
\end{table}

% \begin{table}[t]
%     \begin{center}
%     \centering
%     \resizebox{1.0\columnwidth}{!}{%
%     \begin{tabular}{lcccc}
%     \toprule
%                        & \tf{SST-2} & \tf{TREC} & \tf{SNLI} & \tf{MRPC} \\
%     \midrule
%     Prompt-based FT    & \tf{92.7}   & 84.8      & 77.2      & 74.5   \\
%     \midrule
%     Uniform sampling     & 92.3       & 85.6      & 78.8      & 70.9 \\
%     \tableindent + RoBERTa sel. & \tf{92.7}  & 83.4      & 79.5      & 76.6           \\
%     \tableindent + SBERT sel.   & 92.6       & \tf{87.5} & \tf{79.7} & \tf{77.8}   \\
%     \bottomrule
%     \end{tabular}
%     }
%     \end{center}
% 
%     \caption{
%         Impact of demonstration sampling strategies.
%         Uniform sampling randomly samples demonstrations, while selective (sel.) sampling only takes top sentences measured by the sentence encoders (\S \ref{sec:demonstrations}).
%     }
%     \label{tab:pair_ablation}
% \end{table}

%% file: sections/limitations.tex
%!TEX root = ../main.tex

% \section{Limitations and Future Work}

\section{Discussion}
\label{sec:limitations}

Reformulating NLP tasks as MLM has exciting implications for few-shot learning, but also has limitations. 
First, while LM-BFF greatly outperforms standard fine-tuning, Table~\ref{tab:main_results} shows that, overall, the performance still substantially lags behind 
fine-tuning with thousands of examples, especially for harder tasks. 
Additionally, just like standard fine-tuning, our results also suffer from high variance. 
As described in \S\ref{sec:related_work}, several recent studies have tried to counter instability in few-shot fine-tuning 
%\cite{dodge2020fine, lee2020mixout, zhang2020revisiting} 
and we expect these methods to also help here. 

% Auto search
% We have shown that automatic prompt generation can largely reduce human labor and achieve comparable or even higher results than manual prompts.
With respect to automatic prompt generation, despite its effectiveness, 
we still find it practically challenging to expand the search space, 
or generalize well based on only approximately 32 examples. 
This is partly due to our lingering reliance on \emph{some} manual design---either manual templates (for label word search) or manual label words (for template search), 
%These minimal inputs 
which allows us to get our search off the ground, but does also bias it towards areas of the search space that we might have already imagined.
% Our method also requires either starting from manual templates or label words, so there is still large room for improvement.

% Limitation
Finally, it is important to clarify that LM-BFF favors certain tasks which %, namely those that 
(1) can be naturally posed as a ``fill-in-the-blank'' problem; 
(2) have relatively short input sequences; and 
(3) do not contain many output classes. 
Issues (2) and (3) might be ameliorated with longer-context language models~\cite[e.g.,][]{Beltagy2020Longformer}.
%For tasks such as structured prediction, however, issue (1) is more fundamental, as they are not straightforward to formulate in prompting. We leave it as an open question for future work.
For tasks that are not straightforward to formulate in prompting, such as structured prediction, issue (1) is more fundamental. 
We leave it as an open question for future work.

\begin{figure}[!t]
    \centering
    \includegraphics[width=0.237\textwidth, trim=4 5 0 0, clip]{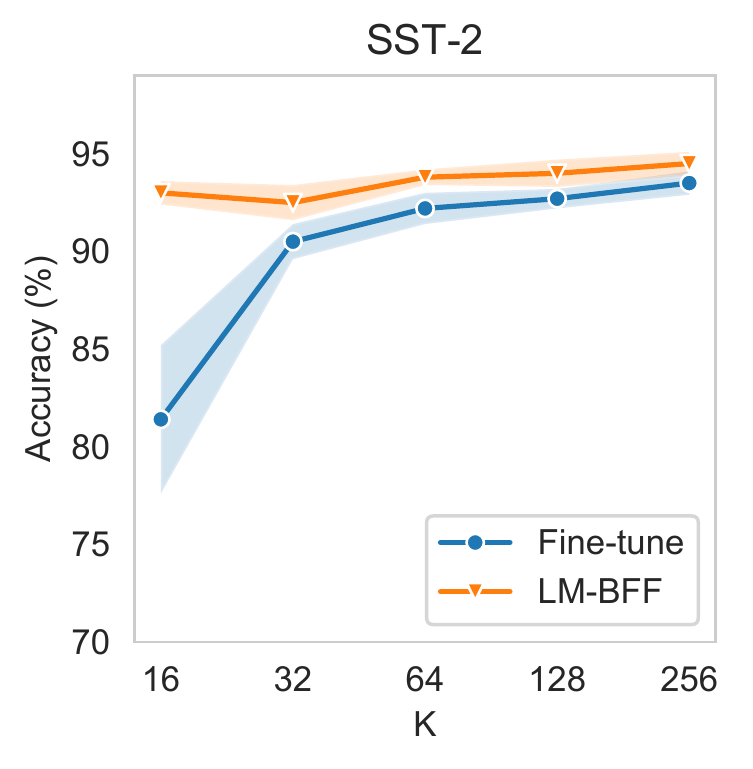}
    \includegraphics[width=0.237\textwidth, trim=4 5 0 0, clip]{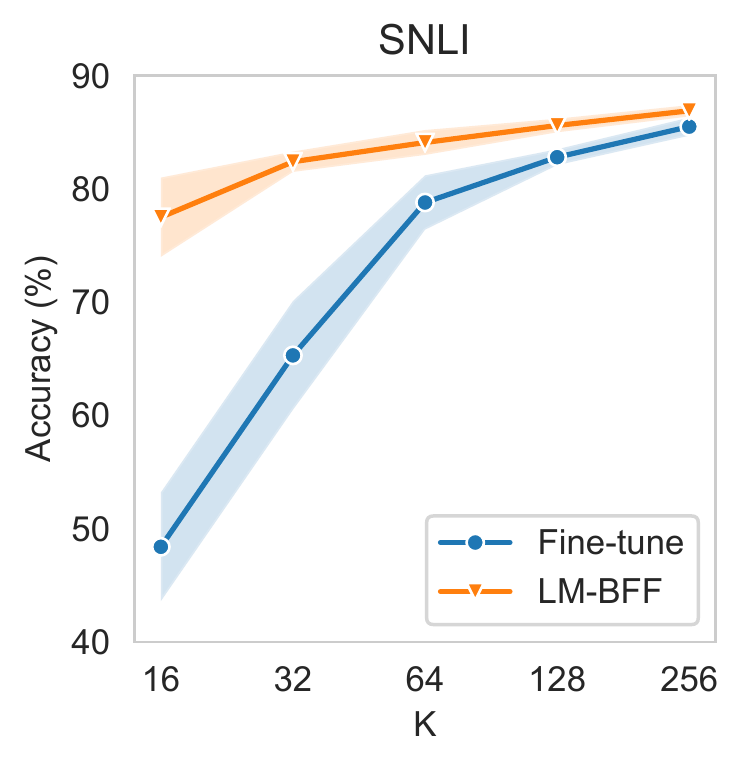}
    \caption{Standard fine-tuning vs our LM-BFF as a function of $K$ (\# instances per class). For lower $K$, our method consistently outperforms standard fine-tuning.}% As $K$ increases the two methods gradually converge.}
    \label{fig:sample}
\end{figure}

%% file: sections/conclusion.tex
%!TEX root = ../main.tex

\section{Conclusion}
\label{sec:conclusion}

In this paper we presented LM-BFF, a set of simple but effective techniques for fine-tuning language models using only a few examples. Our approach proposes to
(1) use prompt-based fine-tuning with automatically searched prompts;
and (2) include selected task demonstrations (training examples) as part of the input context.
We show %through a series of systematic evaluations
that our method outperforms vanilla fine-tuning by up to $30\%$ (and $11$\% on average).
We concluded by discussing the limitations of our approach, and posed open questions for future study.

\section*{Acknowledgements}
We thank the members of Princeton, MIT, Tsinghua NLP groups and the anonymous reviewers for their valuable feedback. TG is supported by a Graduate Fellowship at Princeton University and AF is supported by an NSF Graduate Research Fellowship. This research is also partly supported by a Google Research Scholar Award.

\begin{comment}

\section*{Ethical Considerations}

This work aims to further progress on few-shot learning across a range of NLP tasks. In general, allowing for overall smaller models to be used effectively with few labeled examples expands the accessibility of NLP solutions to more users with limited resources.
Nevertheless, our work makes use of medium-sized pre-trained language models, which---though smaller by design than other contemporary, massive models such as GPT-3---still require care when being deployed in potentially sensitive situations.
More concretely, leveraging the built-in knowledge of pre-trained language models as much as possible is a key feature we exploit in this work for effective few-shot learning.
At the same time, it is well known that these models can inherit the biases present in the data they are trained on---and therefore for some problems (such as those that are race- or gender-sensitive) may require modification and oversight.
However, though our methods may have similar bias concerns as a result, they are agnostic to the specific choice of language models, and we anticipate that developments in ethically-compliant language models can straighforwardly carry through in our use cases.

\end{comment}

%% file: sections/appendix.tex
%!TEX root = ../main.tex

\clearpage
\section{Impact of Development Sets}
\label{app:dev_size}

Table~\ref{tab:exp_large_dev} shows how the size of the development sets can affect the final performance of the model.
For ``No $\ddev$'', we take the same hyper-parameters from~\citet{schick2020exploiting,schick2020size}: batch size = 16, learning rate = 1e-5 and training steps = 250. We also experiment with a variant that we sample a development set of 10 times larger than the training set. We can see that using larger development sets leads to better performance, and this is why we stick to $|\dtrain| = |\ddev|$ in our few-shot setting.

\input{tables/exp_large_dev}

\section{Datasets}
\label{app:datasets}

For SNLI~\cite{bowman2015large_snli} and datasets from GLUE~\cite{wang2019glue}, including SST-2~\cite{socher2013recursive_sst-2}, CoLA~\cite{warstadt2019neural_cola}, MNLI~\cite{williams2018broad_mnli}, QNLI~\cite{rajpurkar2016squad}, RTE~\cite{dagan2005pascal_rte1,bar2006second,giampiccolo2007third_rte3,bentivogli2009fifth_rte4}, MRPC~\cite{dolan2005automatically_mrpc}, QQP\footnote{\url{https://www.quora.com/q/quoradata/}} and STS-B~\cite{cer2017semeval_sts-b}, we follow~\newcite{zhang2020revisiting} and use their original development sets for testing. For datasets which require a cross-validation evaluation---MR~\cite{pang2005seeing_mr}, CR~\cite{hu2004mining_cr}, MPQA~\cite{wiebe2005annotating_mpqa}, Subj~\cite{pang2004sentimental_subj}---we simply randomly sample 2,000 examples as the testing set and leave them out from training. For SST-5~\cite{socher2013recursive_sst-2} and TREC~\cite{voorhees2000building_trec}, we use their official test sets. We show dataset statistics in Table~\ref{tab:datasets}.

\input{tables/datasets.tex}

\section{Experimental Details}
\label{app:exp_details}

\subsection{Hyper-parameter selection}
\label{app:hyper_selection}
For grid search, we take learning rates from \{1e-5, 2e-5, 5e-5\} and batch sizes from \{2, 4, 8\}.
These numbers are picked by pilot experiments on the SST-2 and SNLI datasets.
We also use early stopping to avoid overfitting.
For each trial, we train the model for 1,000 steps, validate the performance every 100 steps, and take the best checkpoint.

\subsection{Prompt-based fine-tuning}
\label{app:prompts}
Table~\ref{tab:manual_prompts} shows all the manual templates and label words we use in experiment.
For automatically template generation, we take the T5-3B\footnote{We take the T5 1.0 checkpoint, which is trained on both unsupervised and downstream task data. We compared it to T5 1.1 (without downstream task data) and did not find a significant difference in generated templates.} model, which is the largest publicly available one that can fit on a single GPU.
For automatically searching label words, we set $k$ to 100 for all tasks except SST-5 and TREC. For SST-5 we set a smaller $k = 30$, as it is a 5-way classification task. For TREC, we observe that filtering $\mathcal{V}^c$ using conditional likelihood alone is still noisy, thus we set $k = 1000$, and then re-rank $\mathcal{V}^c$ by the nearest neighbors of the original manual label words and take the top 30 per class. We set $n$ to 100 in all experiments. 
Due to the large number of trials in automatic search, we take a fixed set of hyper-parameters in this part: batch size of 8 and learning rate of 1e-5.

Since the idea of prompt-based fine-tuning is to make the input and output distribution close to the pre-training, the implementation details are crucial.
For templates, we put extra space before sentences if it is not at the beginning of the input.
Also, we lowercase the first letter of the sentence if it is concatenated with a prefix (e.g., \secondsent~in Table~\ref{tab:manual_prompts}).
Also if one sentence is appended any punctuation (e.g., \firstsent~in Table~\ref{tab:manual_prompts}), then the last character of the original sentence is discarded.
Finally, we prepend a space for label words in $\mapping(\labelset)$. For example, we use ``\_great'' instead of ``great'' in the RoBERTa vocabulary, where ``\_'' stands for space.

\subsection{Fine-tuning with demonstrations}
\label{app:demonstrations}
When using demonstrations, we sample $16$ different sets of demonstrations for each input and average the predicted log probability for each class during inference.
We find that further increasing the number of samples does not bring substantial improvement.
Additional, we have tried different aggregation methods like taking the result with the maximum confidence and we did not find a meaningful improvement.
For selective demonstrations, we take \ttt{roberta-large-nli-stsb- mean-tokens}\footnote{\url{https://github.com/UKPLab/sentence-transformers}}
from \newcite{reimers2019sentence}
as our sentence embedding model.

\section{Comparisons of BERT vs RoBERTa}
\label{app:analysis_bert}

\input{tables/bert.tex}

Table~\ref{tab:bert} compares the results of BERT-large (uncased) and RoBERTa-large in our settings.
Pre-trained BERT provides two segment embeddings (A/B) for different parts of input. The common practice, when fine-tuning BERT, is that using only segment A for single-sentence tasks, and using segment A/B for the two sentences in sentence-pair tasks. 
In our case of incorporating demonstrations, however, we have more than two sentences. Thus we explore the following different strategies for segments:
(1) using the A segment for all sentences (1-seg);
(2) using the A segment for the original input and the B segment for the demonstrations (2-seg); 
%different segment embeddings for the original input and the demonstrations (2-seg);
(3) using different segment embeddings for each sentence ($n$-seg), e.g., for SNLI, we use different segments for each premise and hypothesis in both the original input and the demonstrations, which leads to a total number of 8 segment embeddings. This introduces new segment embeddings (randomly initialized and learned during fine-tuning) as the pre-trained BERT only has two. 

Table~\ref{tab:bert} shows that
prompt-based fine-tuning with demonstrations also works for BERT, and 2-seg works the best when incorporating demonstrations.
Still, we take RoBERTa-large as our main model, for RoBERTa performs much better than BERT and RoBERTa saves the trouble to tune the usage of segment embeddings.

\section{Generated Prompts}
\label{app:generated_prompts}

We demonstrate the top 3 automatically generated templates and label words for all tasks in Table~\ref{tab:full_generated_prompt}. In general, most automatic templates are reasonable and grammatically correct.
For the label words, the generated results look intuitive for most single sentence tasks. For other tasks, the automatic ones can be counterintuitive in some cases.
It is still unclear why the language model picks these words and sometimes they actually work well.
We leave this for future study.

\input{tables/generated_template_full.tex}

% \clearpage

%\clearpage

%% file: tables/exp_large_dev.tex
%!TEX root = ../main.tex

% test, mrpc=f1

\begin{table}[h]
    \begin{center}
    \centering
    \resizebox{1.0\columnwidth}{!}{%
    \begin{tabular}{lcccc}
    \toprule
    {Fine-tuning}         &  \tf{SST-2}  & \tf{SNLI}    &\tf{TREC}   &  \tf{MRPC}     \\
    \midrule
    No $\ddev$             & 79.5    & 49.2  & 83.9  & 77.8    \\
    $|\ddev| = |\dtrain|$    & 81.4     & 48.4   & 88.8   & 76.6    \\
    $|\ddev| = 10|\dtrain|$  & 83.5       & 52.0  & 89.4 & 79.6    \\
    \midrule
    {Prompt-based FT}     & \tf{SST-2}     & \tf{SNLI}  & \tf{TREC}   & \tf{MRPC}     \\
    \midrule
    No $\ddev$             & 92.1      & 75.3 & 84.8   & 70.2    \\
    $|\ddev| = |\dtrain|$    & 92.7      & 77.2  & 84.8  & 74.5     \\
    $|\ddev| = 10|\dtrain|$  & 93.0       & 79.7  & 89.3 & 80.9     \\
    \bottomrule
    \end{tabular}
    }
    \end{center}
    \vspace{-5pt}
    \caption{Impact of different sizes of development sets. Standard deviations are omitted here to save space. For No $|\ddev|$, we use the same set of hyper-parameters as \newcite{schick2020exploiting,schick2020size}.}
    \vspace{-15pt}
    \label{tab:exp_large_dev}
\end{table}

% \begin{table}[h]
%     \begin{center}
%     \centering
%     \resizebox{1.0\columnwidth}{!}{%
%     \begin{tabular}{lcccc}
%     \toprule
%     {Fine-tuning}         &  \tf{SST-2}  & \tf{TREC}   & \tf{SNLI}    & \tf{MRPC}     \\
%     \midrule
%     No $\ddev$             & 79.5   & 83.9  & 49.2   & 77.8    \\
%     $|\ddev| = |\dtrain|$    & 81.4    & 88.8   & 48.4    & 76.6    \\
%     $|\ddev| = 10|\dtrain|$  & 83.5    & 89.4   & 52.0   & 79.6    \\
%     \midrule
%     {Prompt-based FT}     & \tf{SST-2}   & \tf{TREC}   & \tf{SNLI}    & \tf{MRPC}     \\
%     \midrule
%     No $\ddev$             & 92.1    & 84.8   & 75.3   & 70.2    \\
%     $|\ddev| = |\dtrain|$    & 92.7    & 84.8   & 77.2    & 74.5     \\
%     $|\ddev| = 10|\dtrain|$  & 93.0    & 89.3   & 79.7   & 80.9     \\
%     \bottomrule
%     \end{tabular}
%     }
%     \end{center}
%     \vspace{-5pt}
%     \caption{Impact of different sizes of development sets. Standard deviations are omitted here to save space. For No $|\ddev|$, we use the same set of hyper-parameters as \newcite{schick2020exploiting,schick2020size}.}
%     \vspace{-15pt}
%     \label{tab:exp_large_dev}
% \end{table}
% 

%% file: tables/datasets.tex
%!TEX root = ../main.tex

\begin{table*}[t]
\begin{center}
\centering
\resizebox{1.98\columnwidth}{!}{%
\begin{tabular}{llcrrrcl}
\toprule
\tf{Category} & \tf{Dataset} & $|\mathcal{Y}|$ & $L$ & \#Train & \#Test & \tf{Type} & \tf{Labels (classification tasks)} \\
\bottomrule
 & SST-2 & 2 & 19 & 6,920 & 872 & sentiment & positive, negative \\
& SST-5 & 5 & 18 & 8,544 & 2,210 & sentiment & v. pos., positive, neutral, negative, v. neg. \\
& MR & 2 & 20 & 8,662& 2,000 & sentiment & positive, negative \\
single- & CR & 2 & 19 & 1,775 & 2,000 & sentiment & positive, negative \\
sentence & MPQA & 2 & 3 & 8,606 & 2,000 & opinion polarity & positive, negative \\
& Subj & 2 & 23 & 8,000 & 2,000 & subjectivity & subjective, objective \\
& TREC & 6 & 10 & 5,452 & 500 & question cls. & abbr., entity, description, human, loc., num.\\
& CoLA & 2 & 8 & 8,551 & 1,042 & acceptability & grammatical, not\_grammatical\\
\midrule
 & MNLI & 3 & 22/11 & 392,702 & 9,815 & NLI & entailment, neutral, contradiction\\
& SNLI & 3 & 14/8 &  549,367 & 9,842 & NLI & entailment, neutral, contradiction \\
sentence- & QNLI & 2 & 11/30  & 104,743 & 5,463 & NLI & entailment, not\_entailment \\
pair & RTE & 2 &  49/10 & 2,490 & 277 & NLI &  entailment, not\_entailment \\
 & MRPC & 2 & 22/21  & 3,668 & 408 & paraphrase & equivalent, not\_equivalent \\
& QQP & 2 & 12/12 & 363,846 & 40,431 & paraphrase & equivalent, not\_equivalent  \\
& STS-B & $\mathcal{R}$ & 11/11  & 5,749 & 1,500  & sent. similarity & - \\
\bottomrule
\end{tabular}
}
\end{center}
\caption{The datasets evaluated in this work. $|\mathcal{Y}|$: \# of classes for classification tasks (with one exception: STS-B is a real-valued regression task over the interval $[0, 5]$). $L$: average \# of words in input sentence(s). Note that we only sample $\dtrain$ and $\ddev$ of $K \times |\labelset|$ examples from the original training set in our few-shot experiments (\S\ref{sec:setup}).}
\label{tab:datasets}
\end{table*}

%  \danqi{I am debating if we should put the label words or the meaning of labels in this table. }

%% file: tables/bert.tex
%!TEX root = ../main.tex

% Test, mrpc=acc

% test, mrpc=f1
    \begin{table}[t]
        \begin{center}
        \centering
        \resizebox{1.0\columnwidth}{!}{%
        \begin{tabular}{lcccc}
        \toprule
        \tf{BERT-large}        & \tf{SST-2}   & \tf{SNLI}  & \tf{TREC}  & \tf{MRPC} \\
        \midrule
        Fine-tuning              &  79.5  &	51.4    & 80.3& \tf{74.4} \\
        \midrule
        Prompt-based FT        & 85.6        & 59.2   & 79.0      &  66.8 \\
        \tableindent + demo (1-seg)   &  \tf{87.5}  & 50.4 &  77.2  & 68.5 \\
        \tableindent + demo (2-seg)   &  86.1  & \tf{61.3} & 77.9 & 73.2 \\
        \tableindent + demo ($n$-seg)          & 86.4             & 58.6    & \tf{79.6}      & 71.0 \\
        \midrule
        \tf{RoBERTa-large}  & \tf{SST-2}   & \tf{SNLI}      & \tf{TREC} & \tf{MRPC} \\
        \midrule
        Fine-tuning         & 81.4          & 48.4        &  \tf{88.8}       &  76.6\\
        \midrule
        Prompt-based FT     & \tf{92.7}         & 77.2          & 84.8       & 74.5 \\
        \tableindent + demonstrations      & 92.6           & \tf{79.7}     & 87.5          & \tf{77.8} \\
        \bottomrule
        \end{tabular}
        }
        \end{center}
        \caption{A comparison of BERT-large vs RoBERTa-large. We use manual prompts in these experiments.}
        \label{tab:bert}
    \end{table}

%     \begin{table}[t]
%         \begin{center}
%         \centering
%         \resizebox{1.0\columnwidth}{!}{%
%         \begin{tabular}{lcccc}
%         \toprule
%         \tf{BERT-large}        & \tf{SST-2}  & \tf{TREC}   & \tf{SNLI}  & \tf{MRPC} \\
%         \midrule
%         Fine-tuning              &  79.5  & 80.3 &	51.4 & \tf{74.4} \\
%         \midrule
%         Prompt-based FT        & 85.6        & 79.0  & 59.2       &  66.8 \\
%         \tableindent + demo (1-seg)   &  \tf{87.5} &  77.2  & 50.4 & 68.5 \\
%         \tableindent + demo (2-seg)   &  86.1  & 77.9 & \tf{61.3} & 73.2 \\
%         \tableindent + demo ($n$-seg)          & 86.4        & \tf{79.6}        & 58.6       & 71.0 \\
%         \midrule
%         \tf{RoBERTa-large}  & \tf{SST-2}    & \tf{TREC} & \tf{SNLI}     & \tf{MRPC} \\
%         \midrule
%         Fine-tuning         & 81.4          &  \tf{88.8}     & 48.4          &  76.6\\
%         \midrule
%         Prompt-based FT     & \tf{92.7}          & 84.8      & 77.2          & 74.5 \\
%         \tableindent + demonstrations      & 92.6          & 87.5      & \tf{79.7}          & \tf{77.8} \\
%         \bottomrule
%         \end{tabular}
%         }
%         \end{center}
%         \caption{A comparison of BERT-large vs RoBERTa-large. We use manual prompts in these experiments.}
%         \label{tab:bert}
%     \end{table}
%     

%% file: tables/generated_template_full.tex
%!TEX root = ../main.tex

% MRPC/QQP=acc

\begin{table*}[t]
    \begin{center}
    \centering
    \resizebox{1.91\columnwidth}{!}{%
    \begin{tabular}{lll}
    \toprule
    \tf{Task} & \tf{Auto template} & \tf{Auto label words}\\
    \midrule

    \tf{SST-2} & (positive/negative) \\
     & {\sent} A {\mask} one . & irresistible/pathetic \\
     & {\sent} A {\mask} piece . & wonderful/bad \\
     & {\sent} All in all {\mask} . & delicious/bad \\
    \midrule
    \tf{SST-5} & (very positive/positive/neutral/negative/very negative)  \\
     & {\sent} The movie is {\mask} . & wonderful/remarkable/hilarious/better/awful \\
     & {\sent} The music is {\mask} . & wonderful/perfect/hilarious/better/awful \\
     & {\sent} But it is {\mask} . & unforgettable/extraordinary/good/better/terrible \\
    \midrule
    \tf{MR} & (positive/negative) \\
     &  It was {\mask} ! {\sent} & epic/terrible \\
     & {\sent} It's {\mask} . & epic/awful \\
     & {\sent} A {\mask} piece of work . & exquisite/horrible \\
    \midrule
    \tf{CR} & (positive/negative) \\
     & {\sent} It's {\mask} ! & fantastic/horrible \\
     & {\sent} The quality is {\mask} . & neat/pointless \\
     & {\sent} That is {\mask} . & magnificent/unacceptable \\
    \midrule
    \tf{MPQA} & (positive/negative) \\
     & {\sent} is {\mask} . & important/close \\
     & {\sent}, {\mask} ! & needed/bad \\
     & {\sent}. {\mask} . & unexpected/shocking \\
    \midrule
    \tf{Subj} & (subjective/objective) \\
    & {\sent} It's all {\mask} . & everywhere/tragic \\
    & {\sent} It's {\mask} . & everywhere/horrifying \\
    & {\sent} Is it {\mask} ? & something/surreal \\
    \midrule
    \tf{TREC} & (abbreviation/entity/description/human/location/numeric) \\
     &  Q: {\mask} : {\sent} & Application/Advisor/Discussion/Culture/Assignment/Minute \\
     & {\sent} Why {\mask}? & Production/AE/Context/Artist/Assignment/Minute \\
     & {\sent} Answer: {\mask} . & Personality/Advisor/Conclusion/Hum/Assignment/Minute \\
    \midrule
    \tf{CoLA} & (grammatical/not\_grammatical) \\
     & {\sent} You are {\mask} . & one/proof \\
     &  It is {\mask} . {\sent} & wrong/sad \\
     &  I am {\mask} . {\sent} & misleading/disappointing \\
    \midrule
    
        \tf{MNLI} & (entailment/neutral/contradiction) \\
         & {\firstsent} . {\mask} , you are right , {\secondsent} & Fine/Plus/Otherwise \\
         & {\firstsent} . {\mask} you're right {\secondsent} & There/Plus/Otherwise \\
         & {\firstsent} . {\mask} ! {\secondsent} & Meaning/Plus/Otherwise \\
         \midrule
        \tf{SNLI} & (entailment/neutral/contradiction) \\
         & {\firstsent} . {\mask} , no , {\secondsent} & Alright/Watch/Except \\
         & {\firstsent} . {\mask} , in this case {\secondsent} & Hi/Watch/Worse \\
         & {\firstsent} . {\mask} this time {\secondsent} & Regardless/Fortunately/Unless \\
        \midrule
        \tf{QNLI} & (entailment/not\_entailment)  \\
         & {\firstsent} ? {\mask} . Yes , {\secondsent} & Okay/Nonetheless \\
         & {\firstsent} ? {\mask} . It is known that {\secondsent} & Notably/Yet \\
         & {\firstsent} ? {\mask} , however , {\secondsent} & Specifically/Notably \\
        \midrule
        \tf{RTE} & (entailment/not\_entailment) \\
         & {\firstsent} . {\mask} , I believe {\secondsent} & Clearly/Yet \\
         & {\firstsent} . {\mask} , I think that {\secondsent} & Accordingly/meanwhile \\
         & {\firstsent} . {\mask} , I think {\secondsent} & So/Meanwhile \\
        \midrule
        \tf{MRPC} & (equivalent/not\_equivalent) \\
         & {\firstsent} . {\mask} ! {\secondsent} & Rather/Alas \\
         & {\firstsent} . {\mask} . This is the first time {\secondsent} & At/Thus \\
         & {\firstsent} . {\mask} . That's right . {\secondsent} & Instead/Moreover \\
        \midrule
        \tf{QQP} & (equivalent/not\_equivalent)\\
         & {\firstsent} ? {\mask} , but {\secondsent} & Me/Since \\
         & {\firstsent} ? {\mask} , please , {\secondsent} & Um/Best \\
         & {\firstsent} ? {\mask} , I want to know {\secondsent} & Ironically/Beyond \\
        \midrule
        \tf{STS-B} & ($y_u$/$y_l$)\\
         & {\firstsent} . {\mask} sir {\secondsent} & Note/Next \\
         & {\firstsent} . {\mask} , it is not . {\secondsent} & Yesterday/meanwhile \\
         & {\firstsent} . {\mask} . It is {\secondsent} & Yeah/meanwhile \\
        \bottomrule
    
    \end{tabular}
    }
    \end{center}
    \caption{Top 3 automatically generated templates and label words for all tasks based on one split of $K=16$ training examples. Note that automatic template results are based on manual label words and automatic label word results are based on manual templates provided in Table~\ref{tab:manual_prompts}.}
    \label{tab:full_generated_prompt}
    \end{table*}

%% file: main.bbl
\begin{thebibliography}{50}
\expandafter\ifx\csname natexlab\endcsname\relax\def\natexlab#1{#1}\fi

\bibitem[{Bansal et~al.(2020{\natexlab{a}})Bansal, Jha, and
  McCallum}]{bansal2020learning}
Trapit Bansal, Rishikesh Jha, and Andrew McCallum. 2020{\natexlab{a}}.
\newblock Learning to few-shot learn across diverse natural language
  classification tasks.
\newblock In \emph{International Conference on Computational Linguistics
  (COLING)}.

\bibitem[{Bansal et~al.(2020{\natexlab{b}})Bansal, Jha, Munkhdalai, and
  McCallum}]{bansal2020self}
Trapit Bansal, Rishikesh Jha, Tsendsuren Munkhdalai, and Andrew McCallum.
  2020{\natexlab{b}}.
\newblock Self-supervised meta-learning for few-shot natural language
  classification tasks.
\newblock In \emph{Empirical Methods in Natural Language Processing (EMNLP)}.

\bibitem[{Bao et~al.(2020)Bao, Wu, Chang, and Barzilay}]{bao2020fewshot}
Yujia Bao, Menghua Wu, Shiyu Chang, and Regina Barzilay. 2020.
\newblock Few-shot text classification with distributional signatures.
\newblock In \emph{International Conference on Learning Representations
  (ICLR)}.

\bibitem[{Bar~Haim et~al.(2006)Bar~Haim, Dagan, Dolan, Ferro, Giampiccolo,
  Magnini, and Szpektor}]{bar2006second}
Roy Bar~Haim, Ido Dagan, Bill Dolan, Lisa Ferro, Danilo Giampiccolo, Bernardo
  Magnini, and Idan Szpektor. 2006.
\newblock The second {PASCAL} recognising textual entailment challenge.

\bibitem[{Beltagy et~al.(2020)Beltagy, Peters, and
  Cohan}]{Beltagy2020Longformer}
Iz~Beltagy, Matthew~E. Peters, and Arman Cohan. 2020.
\newblock Longformer: The long-document {Transformer}.
\newblock \emph{arXiv:2004.05150}.

\bibitem[{Bentivogli et~al.(2009)Bentivogli, Clark, Dagan, and
  Giampiccolo}]{bentivogli2009fifth_rte4}
Luisa Bentivogli, Peter Clark, Ido Dagan, and Danilo Giampiccolo. 2009.
\newblock The fifth {PASCAL} recognizing textual entailment challenge.
\newblock In \emph{TAC}.

\bibitem[{Bowman et~al.(2015)Bowman, Angeli, Potts, and
  Manning}]{bowman2015large_snli}
Samuel Bowman, Gabor Angeli, Christopher Potts, and Christopher~D Manning.
  2015.
\newblock A large annotated corpus for learning natural language inference.
\newblock In \emph{Empirical Methods in Natural Language Processing (EMNLP)}.

\bibitem[{Brown et~al.(2020)Brown, Mann, Ryder, Subbiah, Kaplan, Dhariwal,
  Neelakantan, Shyam, Sastry, Askell et~al.}]{brown2020language}
Tom~B Brown, Benjamin Mann, Nick Ryder, Melanie Subbiah, Jared Kaplan, Prafulla
  Dhariwal, Arvind Neelakantan, Pranav Shyam, Girish Sastry, Amanda Askell,
  et~al. 2020.
\newblock Language models are few-shot learners.
\newblock In \emph{Advances in Neural Information Processing Systems
  (NeurIPS)}.

\bibitem[{Cer et~al.(2017)Cer, Diab, Agirre, Lopez-Gazpio, and
  Specia}]{cer2017semeval_sts-b}
Daniel Cer, Mona Diab, Eneko Agirre, I{\~n}igo Lopez-Gazpio, and Lucia Specia.
  2017.
\newblock {S}em{E}val-2017 task 1: Semantic textual similarity multilingual and
  crosslingual focused evaluation.
\newblock In \emph{the 11th International Workshop on Semantic Evaluation
  ({S}em{E}val-2017)}.

\bibitem[{Chen et~al.(2020)Chen, Yang, and Yang}]{chen2020mixtext}
Jiaao Chen, Zichao Yang, and Diyi Yang. 2020.
\newblock {MixText}: Linguistically-informed interpolation of hidden space for
  semi-supervised text classification.
\newblock In \emph{Association for Computational Linguistics (ACL)}.

\bibitem[{Dagan et~al.(2005)Dagan, Glickman, and
  Magnini}]{dagan2005pascal_rte1}
Ido Dagan, Oren Glickman, and Bernardo Magnini. 2005.
\newblock The {PASCAL} recognising textual entailment challenge.
\newblock In \emph{the First International Conference on Machine Learning
  Challenges: Evaluating Predictive Uncertainty Visual Object Classification,
  and Recognizing Textual Entailment}.

\bibitem[{Davison et~al.(2019)Davison, Feldman, and
  Rush}]{davison2019commonsense}
Joe Davison, Joshua Feldman, and Alexander~M Rush. 2019.
\newblock Commonsense knowledge mining from pretrained models.
\newblock In \emph{Empirical Methods in Natural Language Processing (EMNLP)}.

\bibitem[{Devlin et~al.(2019)Devlin, Chang, Lee, and
  Toutanova}]{devlin2019bert}
Jacob Devlin, Ming-Wei Chang, Kenton Lee, and Kristina Toutanova. 2019.
\newblock {BERT}: Pre-training of deep bidirectional {Transformers} for
  language understanding.
\newblock In \emph{North American Chapter of the Association for Computational
  Linguistics (NAACL)}.

\bibitem[{Dodge et~al.(2020)Dodge, Ilharco, Schwartz, Farhadi, Hajishirzi, and
  Smith}]{dodge2020fine}
Jesse Dodge, Gabriel Ilharco, Roy Schwartz, Ali Farhadi, Hannaneh Hajishirzi,
  and Noah Smith. 2020.
\newblock Fine-tuning pretrained language models: Weight initializations, data
  orders, and early stopping.
\newblock \emph{arXiv preprint arXiv:2002.06305}.

\bibitem[{Dolan and Brockett(2005)}]{dolan2005automatically_mrpc}
William~B. Dolan and Chris Brockett. 2005.
\newblock Automatically constructing a corpus of sentential paraphrases.
\newblock In \emph{the Third International Workshop on Paraphrasing
  ({IWP}2005)}.

\bibitem[{Giampiccolo et~al.(2007)Giampiccolo, Magnini, Dagan, and
  Dolan}]{giampiccolo2007third_rte3}
Danilo Giampiccolo, Bernardo Magnini, Ido Dagan, and Bill Dolan. 2007.
\newblock The third {PASCAL} recognizing textual entailment challenge.
\newblock In \emph{the {ACL}-{PASCAL} Workshop on Textual Entailment and
  Paraphrasing}.

\bibitem[{Han et~al.(2018)Han, Zhu, Yu, Wang, Yao, Liu, and
  Sun}]{han2018fewrel}
Xu~Han, Hao Zhu, Pengfei Yu, Ziyun Wang, Yuan Yao, Zhiyuan Liu, and Maosong
  Sun. 2018.
\newblock Fewrel: A large-scale supervised few-shot relation classification
  dataset with state-of-the-art evaluation.
\newblock In \emph{Empirical Methods in Natural Language Processing (EMNLP)}.

\bibitem[{Howard and Ruder(2018)}]{howard2018universal}
Jeremy Howard and Sebastian Ruder. 2018.
\newblock Universal language model fine-tuning for text classification.
\newblock In \emph{Association for Computational Linguistics (ACL)}.

\bibitem[{Hu and Liu(2004)}]{hu2004mining_cr}
Minqing Hu and Bing Liu. 2004.
\newblock Mining and summarizing customer reviews.
\newblock In \emph{ACM SIGKDD international conference on Knowledge discovery
  and data mining}.

\bibitem[{Jiang et~al.(2020)Jiang, Xu, Araki, and Neubig}]{jiang2020can}
Zhengbao Jiang, Frank~F Xu, Jun Araki, and Graham Neubig. 2020.
\newblock How can we know what language models know?
\newblock \emph{Transactions of the Association of Computational Linguistics
  (TACL)}.

\bibitem[{Lee et~al.(2020)Lee, Cho, and Kang}]{lee2020mixout}
Cheolhyoung Lee, Kyunghyun Cho, and Wanmo Kang. 2020.
\newblock Mixout: Effective regularization to finetune large-scale pretrained
  language models.
\newblock In \emph{International Conference on Learning Representations
  (ICLR)}.

\bibitem[{Liu et~al.(2019)Liu, Ott, Goyal, Du, Joshi, Chen, Levy, Lewis,
  Zettlemoyer, and Stoyanov}]{liu2019roberta}
Yinhan Liu, Myle Ott, Naman Goyal, Jingfei Du, Mandar Joshi, Danqi Chen, Omer
  Levy, Mike Lewis, Luke Zettlemoyer, and Veselin Stoyanov. 2019.
\newblock {RoBERTa}: {A} robustly optimized {BERT} pretraining approach.
\newblock \emph{arXiv preprint arXiv:1907.11692}.

\bibitem[{Mettes et~al.(2019)Mettes, van~der Pol, and
  Snoek}]{mettes2019hyperspherical}
Pascal Mettes, Elise van~der Pol, and Cees Snoek. 2019.
\newblock Hyperspherical prototype networks.
\newblock In \emph{Advances in Neural Information Processing Systems
  (NeurIPS)}.

\bibitem[{Miyato et~al.(2017)Miyato, Dai, and
  Goodfellow}]{miyato2017adversarial}
Takeru Miyato, Andrew~M Dai, and Ian Goodfellow. 2017.
\newblock Adversarial training methods for semi-supervised text classification.
\newblock In \emph{International Conference on Learning Representations
  (ICLR)}.

\bibitem[{Pang and Lee(2004)}]{pang2004sentimental_subj}
Bo~Pang and Lillian Lee. 2004.
\newblock A sentimental education: Sentiment analysis using subjectivity
  summarization based on minimum cuts.
\newblock In \emph{Association for Computational Linguistics (ACL)}.

\bibitem[{Pang and Lee(2005)}]{pang2005seeing_mr}
Bo~Pang and Lillian Lee. 2005.
\newblock Seeing stars: Exploiting class relationships for sentiment
  categorization with respect to rating scales.
\newblock In \emph{Association for Computational Linguistics (ACL)}.

\bibitem[{Petroni et~al.(2019)Petroni, Rockt{\"a}schel, Riedel, Lewis, Bakhtin,
  Wu, and Miller}]{petroni2019language}
Fabio Petroni, Tim Rockt{\"a}schel, Sebastian Riedel, Patrick Lewis, Anton
  Bakhtin, Yuxiang Wu, and Alexander Miller. 2019.
\newblock Language models as knowledge bases?
\newblock In \emph{Empirical Methods in Natural Language Processing (EMNLP)}.

\bibitem[{Phang et~al.(2018)Phang, F{\'e}vry, and Bowman}]{phang2018sentence}
Jason Phang, Thibault F{\'e}vry, and Samuel~R Bowman. 2018.
\newblock Sentence encoders on {STILTs}: Supplementary training on intermediate
  labeled-data tasks.
\newblock \emph{arXiv preprint arXiv:1811.01088}.

\bibitem[{Radford et~al.(2018)Radford, Narasimhan, Salimans, and
  Sutskever}]{radford2018improving}
Alec Radford, Karthik Narasimhan, Tim Salimans, and Ilya Sutskever. 2018.
\newblock Improving language understanding by generative pre-training.
\newblock Technical report, OpenAI.

\bibitem[{Radford et~al.(2019)Radford, Wu, Child, Luan, Amodei, and
  Sutskever}]{radford2019language}
Alec Radford, Jeff Wu, Rewon Child, David Luan, Dario Amodei, and Ilya
  Sutskever. 2019.
\newblock Language models are unsupervised multitask learners.
\newblock Technical report, OpenAI.

\bibitem[{Raffel et~al.(2020)Raffel, Shazeer, Roberts, Lee, Narang, Matena,
  Zhou, Li, and Liu}]{raffel2020exploring}
Colin Raffel, Noam Shazeer, Adam Roberts, Katherine Lee, Sharan Narang, Michael
  Matena, Yanqi Zhou, Wei Li, and Peter~J Liu. 2020.
\newblock Exploring the limits of transfer learning with a unified text-to-text
  {Transformer}.
\newblock \emph{The Journal of Machine Learning Research (JMLR)}, 21(140).

\bibitem[{Rajpurkar et~al.(2016)Rajpurkar, Zhang, Lopyrev, and
  Liang}]{rajpurkar2016squad}
Pranav Rajpurkar, Jian Zhang, Konstantin Lopyrev, and Percy Liang. 2016.
\newblock {SQ}u{AD}: 100,000+ questions for machine comprehension of text.
\newblock In \emph{Empirical Methods in Natural Language Processing (EMNLP)}.

\bibitem[{Reimers and Gurevych(2019)}]{reimers2019sentence}
Nils Reimers and Iryna Gurevych. 2019.
\newblock {Sentence-BERT}: Sentence embeddings using {Siamese} {BERT}-networks.
\newblock In \emph{Empirical Methods in Natural Language Processing and
  International Joint Conference on Natural Language Processing
  (EMNLP-IJCNLP)}.

\bibitem[{Schick et~al.(2020)Schick, Schmid, and
  Sch{\"u}tze}]{schick2020automatically}
Timo Schick, Helmut Schmid, and Hinrich Sch{\"u}tze. 2020.
\newblock Automatically identifying words that can serve as labels for few-shot
  text classification.
\newblock In \emph{International Conference on Computational Linguistics
  (COLING)}.

\bibitem[{Schick and Sch{\"u}tze(2021{\natexlab{a}})}]{schick2020exploiting}
Timo Schick and Hinrich Sch{\"u}tze. 2021{\natexlab{a}}.
\newblock Exploiting cloze questions for few-shot text classification and
  natural language inference.
\newblock In \emph{European Chapter of the Association for Computational
  Linguistics (EACL)}.

\bibitem[{Schick and Sch{\"u}tze(2021{\natexlab{b}})}]{schick2020size}
Timo Schick and Hinrich Sch{\"u}tze. 2021{\natexlab{b}}.
\newblock It's not just size that matters: Small language models are also
  few-shot learners.
\newblock In \emph{North American Chapter of the Association for Computational
  Linguistics (NAACL)}.

\bibitem[{Shin et~al.(2020)Shin, Razeghi, IV, Wallace, and
  Singh}]{shin2020autoprompt}
Taylor Shin, Yasaman Razeghi, Robert L.~Logan IV, Eric Wallace, and Sameer
  Singh. 2020.
\newblock {AutoPrompt}: Automatic prompt construction for masked language
  models.
\newblock In \emph{Empirical Methods in Natural Language Processing (EMNLP)}.

\bibitem[{Socher et~al.(2013)Socher, Perelygin, Wu, Chuang, Manning, Ng, and
  Potts}]{socher2013recursive_sst-2}
Richard Socher, Alex Perelygin, Jean Wu, Jason Chuang, Christopher~D. Manning,
  Andrew Ng, and Christopher Potts. 2013.
\newblock Recursive deep models for semantic compositionality over a sentiment
  treebank.
\newblock In \emph{Empirical Methods in Natural Language Processing (EMNLP)}.

\bibitem[{Talmor et~al.(2020)Talmor, Elazar, Goldberg, and
  Berant}]{talmor2020olmpics}
Alon Talmor, Yanai Elazar, Yoav Goldberg, and Jonathan Berant. 2020.
\newblock {oLMpics}-on what language model pre-training captures.
\newblock \emph{Transactions of the Association of Computational Linguistics
  (TACL)}, 8.

\bibitem[{Trinh and Le(2018)}]{trinh2018simple}
Trieu~H Trinh and Quoc~V Le. 2018.
\newblock A simple method for commonsense reasoning.
\newblock \emph{arXiv preprint arXiv:1806.02847}.

\bibitem[{Voorhees and Tice(2000)}]{voorhees2000building_trec}
Ellen~M Voorhees and Dawn~M Tice. 2000.
\newblock Building a question answering test collection.
\newblock In \emph{the 23rd annual international ACM SIGIR conference on
  Research and development in information retrieval}.

\bibitem[{Wang et~al.(2019)Wang, Singh, Michael, Hill, Levy, and
  Bowman}]{wang2019glue}
Alex Wang, Amanpreet Singh, Julian Michael, Felix Hill, Omer Levy, and Samuel~R
  Bowman. 2019.
\newblock {GLUE}: A multi-task benchmark and analysis platform for natural
  language understanding.
\newblock In \emph{International Conference on Learning Representations
  (ICLR)}.

\bibitem[{Warstadt et~al.(2019)Warstadt, Singh, and
  Bowman}]{warstadt2019neural_cola}
Alex Warstadt, Amanpreet Singh, and Samuel~R. Bowman. 2019.
\newblock Neural network acceptability judgments.
\newblock \emph{Transactions of the Association of Computational Linguistics
  (TACL)}, 7.

\bibitem[{Wiebe et~al.(2005)Wiebe, Wilson, and
  Cardie}]{wiebe2005annotating_mpqa}
Janyce Wiebe, Theresa Wilson, and Claire Cardie. 2005.
\newblock Annotating expressions of opinions and emotions in language.
\newblock \emph{Language resources and evaluation}, 39(2-3).

\bibitem[{Williams et~al.(2018)Williams, Nangia, and
  Bowman}]{williams2018broad_mnli}
Adina Williams, Nikita Nangia, and Samuel Bowman. 2018.
\newblock A broad-coverage challenge corpus for sentence understanding through
  inference.
\newblock In \emph{North American Chapter of the Association for Computational
  Linguistics: Human Language Technologies (NAACL-HLT)}.

\bibitem[{Xie et~al.(2020)Xie, Dai, Hovy, Luong, and Le}]{xie2020unsupervised}
Qizhe Xie, Zihang Dai, Eduard Hovy, Thang Luong, and Quoc Le. 2020.
\newblock Unsupervised data augmentation for consistency training.
\newblock \emph{Advances in Neural Information Processing Systems (NeurIPS)},
  33.

\bibitem[{Yin et~al.(2020)Yin, Rajani, Radev, Socher, and
  Xiong}]{yin2020universal}
Wenpeng Yin, Nazneen~Fatema Rajani, Dragomir Radev, Richard Socher, and Caiming
  Xiong. 2020.
\newblock Universal natural language processing with limited annotations: Try
  few-shot textual entailment as a start.
\newblock In \emph{Empirical Methods in Natural Language Processing (EMNLP)}.

\bibitem[{Yu et~al.(2018)Yu, Guo, Yi, Chang, Potdar, Cheng, Tesauro, Wang, and
  Zhou}]{yu2018diverse}
Mo~Yu, Xiaoxiao Guo, Jinfeng Yi, Shiyu Chang, Saloni Potdar, Yu~Cheng, Gerald
  Tesauro, Haoyu Wang, and Bowen Zhou. 2018.
\newblock Diverse few-shot text classification with multiple metrics.
\newblock In \emph{North American Chapter of the Association for Computational
  Linguistics (NAACL)}.

\bibitem[{Zhang et~al.(2021)Zhang, Wu, Katiyar, Weinberger, and
  Artzi}]{zhang2020revisiting}
Tianyi Zhang, Felix Wu, Arzoo Katiyar, Kilian~Q Weinberger, and Yoav Artzi.
  2021.
\newblock Revisiting few-sample {BERT} fine-tuning.
\newblock In \emph{International Conference on Learning Representations
  (ICLR)}.

\bibitem[{Zhong et~al.(2021)Zhong, Friedman, and Chen}]{zhong2021factual}
Zexuan Zhong, Dan Friedman, and Danqi Chen. 2021.
\newblock Factual probing is {[MASK]}: Learning vs. learning to recall.
\newblock In \emph{North American Association for Computational Linguistics
  (NAACL)}.

\end{thebibliography}
